\documentclass[%
 reprint]{revtex4-1}

\usepackage{hyperref}
\usepackage{booktabs}
\usepackage{times}
\usepackage{graphicx}
\usepackage{dcolumn} %
\usepackage{bm}      %
\usepackage{fontenc}
\usepackage{rotating}
\usepackage[dvipsnames]{xcolor}
\usepackage{epstopdf}

\usepackage{comment}

\usepackage{amsthm}
\usepackage{amssymb}
\usepackage{amsfonts}
\usepackage{amsmath}
\usepackage{mathrsfs}
\usepackage{mathtools}

\usepackage{graphicx}
\graphicspath{{./figures/}}          
\DeclareGraphicsExtensions{.pdf,.jpeg,.png,.jpg,.eps} 
\usepackage{wrapfig}
\usepackage[caption=false]{subfig}

\usepackage{soul} 

\usepackage{xr}
\usepackage[toc,page]{appendix}

\DeclareMathOperator*{\argmin}{arg\,min}

\begin{document}

\title{Tailoring Echo State Networks for Optimal Learning}

\affiliation{Max Planck Institute for Mathematics in the Sciences, 04103 Leipzig, Germany}
\affiliation{School of Physics Science and Engineering, Tongji University, 200092 Shanghai, China }
\affiliation{Channing Division of Network Medicine, Brigham and Women's Hospital, Harvard Medical School, Boston, Massachusetts 02115, USA}
\affiliation{Center for Cancer Systems Biology, Dana Farber Cancer Institute, Boston, Massachusetts 02115, USA}

\author{Pau Vilimelis Aceituno}
\affiliation{Channing Division of Network Medicine, Brigham and Women's Hospital, Harvard Medical School, Boston, Massachusetts 02115, USA}
\affiliation{Max Planck Institute for Mathematics in the Sciences, 04103 Leipzig, Germany}
\author{Gang Yan}
\affiliation{School of Physics Science and Engineering, Tongji University, 200092 Shanghai, China }
\author{ Yang-Yu Liu}
\affiliation{Channing Division of Network Medicine, Brigham and Women's Hospital, Harvard Medical School, Boston, Massachusetts 02115, USA}
\affiliation{Center for Cancer Systems Biology, Dana Farber Cancer Institute, Boston, Massachusetts 02115, USA}

\begin{abstract}
As one of the most important paradigms of recurrent neural networks, the echo state network (ESN) has been applied to a wide range of fields, from robotics to medicine, finance, and language processing. A key feature of the ESN paradigm is its reservoir --- a directed and weighted network of neurons that projects the input time series into a high dimensional space where linear regression or classification can be applied. Despite extensive studies, the impact of the reservoir network on the ESN performance remains unclear. By analyzing the dynamics of the reservoir we show that the ensemble of eigenvalues of the network contribute to the ESN memory capacity. Moreover, we find that adding short loops to the reservoir network can tailor ESN for specific tasks and optimize learning. We validate our findings by applying ESN to forecast both synthetic and real benchmark time series. Our results provide a new way to design task-specific ESN. More importantly, it demonstrates the power of combining tools from physics, dynamical systems and network science to offer new insights in recurrent neural networks.

\end{abstract}

\pacs{Valid PACS appear here}
\maketitle


	\section{Introduction}
Echo state network (ESN) is a promising paradigm of recurrent neural networks that can be used to model and predict the temporal behavior of nonlinear dynamic systems \cite{jaeger2004harnessing}. As a special form of recurrent neural networks, ESN has feedback loops in the randomly assigned and fixed synaptic connections and trains only a linear combination of the neurons' states. This fundamentally differs from the traditional feed-forward neural networks, which have multiple layers but no cycles \cite{christopher2006pattern} and simplifies other recurrent neural network architectures that suffer from the difficulty in training synaptic connections \cite{pascanu2013difficulty}. Owing to its simplicity, flexibility and empirical success, ESN and its variants have attracted intense interest during the last decade \cite{jaeger2002tutorial,jaeger2001echo}, and have been applied to many different tasks such as electric load forecasting\cite{deihimi2012application}, robotic control\cite{ploger2003echo}, epilepsy forecasting\cite{buteneers2008real}, stock price prediction\cite{lin2009short}, grammar processing \cite{tong2007learning}, and many others \cite{verplancke2010novel,newton2012neurally,coulibaly2010reservoir,pathak2018model}.

An ESN can be viewed as a dynamic system from which the information of input signals is extracted \cite{dambre2012information}. It has been shown that the information processing capacity of a dynamic system, in theory, depends only on the number of linearly independent variables or, in our case, neurons \cite{dambre2012information,duport2012all,maass2002real}. Yet, the theoretical capacity does not imply that all implementations are practical \cite{whitley2005complexity,busing2010connectivity}, nor does it mean that any reservoir is equally desirable for a given task. A clear example is the effect of the reservoir's spectral radius (i.e., the largest eigenvalue in modulus): an ESN with a larger spectral radius has longer-lasting memory, indicating that it can better process information from past inputs \cite{jaeger2002tutorial}.

Over the last decade, a plethora of studies have focused on finding good reservoir networks. Those studies fall broadly into two categories. First, for specific tasks, systematical parameter searches provide some improvement over classical Monte Carlo reservoir selection \cite{ferreira2011comparing, jiang2008supervised, deng2006complex,liebald2004exploration,rodriguez2017optimal}, but remain costly and do not offer a significant performance improvement or better understanding. Second, some authors have explored networks with some particular characteristics that make them desirable, typically with long memory \cite{farkavs2016computational, rodan2012simple,strauss2012design} or ``rich" dynamics \cite{ozturk2007analysis,boedecker2012information}, although the desirability of those traits typically are task-specific. Here we focus on a ``mechanistic" understanding of reservoir dynamics, but instead of trying to find reservoirs with predefined features, we design reservoirs whose dynamics are tailored to specific problems.

We start by showing how the correlations between neurons define the memory of ESN, and demonstrate that those correlations are determined by the eigenvalues of the reservoir's adjacency matrix. This result allows us to easily assess the memory capacity of a particular reservoir network, unifying previous results\cite{farkavs2016computational,jaeger2001short,strauss2012design,rodan2012simple}. Then we go beyond the current ESN practice and reveal previously unexplored optimization strategies. In particular, we show that adding short loops to the reservoir network can create resonant frequencies and enhance ESN performance by adapting the reservoir to specific tasks. Our results provide insights into the memory capacity of dynamic systems, offering potential improvements to other types of artificial neural networks.

\section{The ESN Framework}

\begin{figure}[h!]
	\centering
	\includegraphics[width=0.5\textwidth]{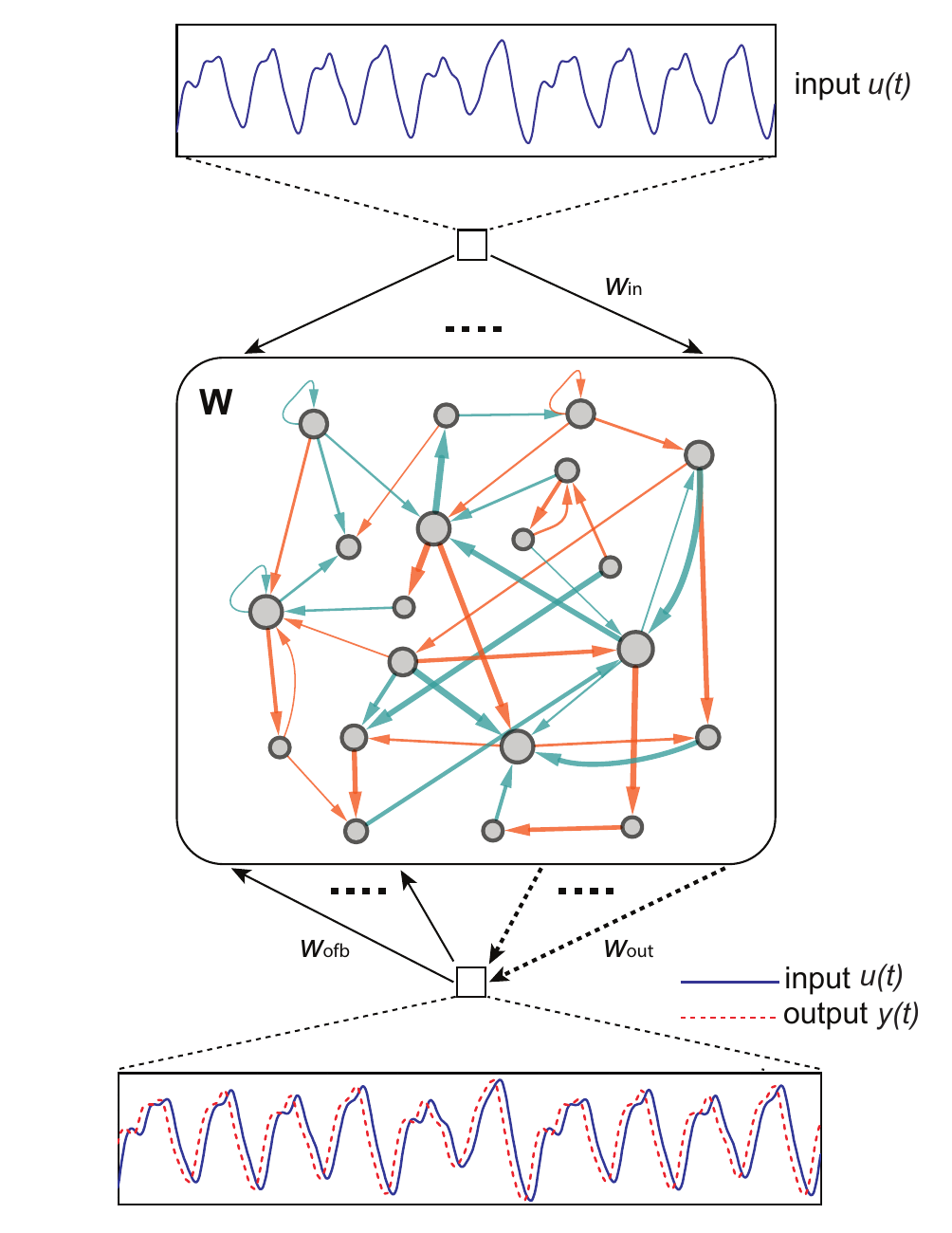}\\
	\caption{\noindent \textbf{The basic schema of an ESN.} The input signal $u(t)$ goes to each neuron in the reservoir with input weights $\text{w}_{\text{in}}$, the neurons send their states to their neighbors according to the matrix $\mathbf{W}$, and the contribution of each neuron to the output $y(t)$ is collected by $\text{w}_{\text{out}}$. 
		\label{fig:ESNschema}}
\end{figure}

The basic ESN architecture is depicted in Fig. \ref{fig:ESNschema}. With different coefficients (weights), the input signal and the predicted output from the previous time step are sent to all neurons in the reservoir. The output is calculated as a linear combination of the neuron states and the input. At each time step, each neuron updates its state according to the current input it receives, the output prediction and its neighboring neurons' states from the previous time step. Formally, the discrete-time dynamics of an ESN with $N$ neurons, one input and one output is governed by 
\begin{equation}\label{eq:ESNModel}
\begin{aligned}
\mathbf{x}(t) &= f(\mathbf{W}\mathbf{x}(t-1) + \mathbf{w}_{\text{in}}u(t) + \mathbf{w}_{\text{ofb}} y(t-1)),
\end{aligned}
\end{equation} 
\begin{equation}\label{eq:ESNReadout}
\begin{aligned}
y(t) &= \mathbf{w}_{\text{out}}\begin{pmatrix}\mathbf{x}(t)\\u(t)\end{pmatrix},
\end{aligned}
\end{equation}
where $\mathbf{x}(t) = [x_1(t),x_2(t),\ldots,x_N(t)]^{\top} \in \mathbb{R}^N$ denotes the state of the $N$ neurons at time $t$, $u(t)\in \mathbb{R}$ is the input signal, the vector $\begin{pmatrix}{\bf x}(t) \\[-0.1cm]u(t)\end{pmatrix}\in \mathbb{R}^{N+1}$ represents the concatenation of $\mathbf{x}(t)$ and $u(t)$, and $y(t)\in \mathbb{R}$ is the output at time $t$. There are various possibilities for the nonlinear function $f$, the most common ones being the logistic sigmoid and the hyperbolic tangent\cite{christopher2006pattern}. Without loss of generality we choose the latter in this work. The matrix $\mathbf{W}\in \mathbb{R}^{N \times N}$ is the weighted adjacency matrix of the reservoir network describing the fixed wiring diagram of $N$ neurons in the reservoir. There is a rich literature on the conditions that the matrix $\mathbf{W}$ must fulfill \cite{jaeger2007discovering,yildiz2012re,buehner2006tighter,manjunath2012theory}. Here we adopt a conservative and simple condition that the reservoir must be a stable dynamic system. The vector $\mathbf{w}_{\text{in}}\in \mathbb{R}^{N}$ captures the fixed weights of the input connections, which we draw from a uniform distribution in the interval $[-1,1]$. The vector $\mathbf{w}_{\text{ofb}}\in \mathbb{R}^{N}$ denotes the fixed weights of the feedback connections from the output to the $N$ neurons, which can induce instabilities if chosen carelessly and may be zero in some tasks \cite{jaeger2002tutorial}. Finally, the row vector $\mathbf{w}_{\text{out}}\in \mathbb{R}^{1 \times (N+1)}$ represents the trainable weights of the readout connections from the $N$ neurons and the input to the output.

A key feature of ESN is that $\mathbf{W}$, $\mathbf{w}_{\text{in}}$ and $\mathbf{w}_{\text{ofb}}$ are all predetermined before the training process, and only the weights of the readout connections $\mathbf{w}_{\text{out}}$ are modified to $\mathbf{w}^{*}_{\text{out}}$ during the training process:
\begin{equation}\label{eq:ESNRegression}
\mathbf{w}^{*}_{\text{out}} =\argmin_{\mathbf{w}_{\text{out}}}\sum_{t=t_0}^{t_0+T}(y(t)-\hat{y}(t))^2,
\end{equation}
where $t_0$ is the starting time, $T$ is the interval of the training, and $\hat{y}(t)$ is the target output obtained from the training data. In other words, $\mathbf{w}^{*}_{\text{out}}$ is the linear regression weights of the desired output $\hat{y}(t)$ on the extended state vector $\begin{pmatrix}{\bf x}(t) \\[-0.1cm]u(t)\end{pmatrix}$, which can be easily solved (see SI Sec. I. for details). Hence, $\mathbf{w}^{*}_{\text{out}}$ captures the underlying mechanism of the dynamic system that produces the training data. Indeed, the right choice of $\mathbf{w}^{*}_{\text{out}}$ can be used to forecast, reconstruct or filter nonlinear time series.

Note that there is a rich literature on methods to improve the ESN performance such as using regularization in the computation of $\mathbf{w}^{*}_{\text{out}}$ \cite{jaeger2002tutorial}, controlling the input weights \cite{strauss2012design} or changing the dynamics of the neurons\cite{lukovsevivcius2009reservoir}. Those results, while relevant and important for applications, are tangential to our study. Therefore in this work we will use the simplest version of the ESN as presented above.

\section{Results}

\begin{figure*}[ht!]
	\centering
	\includegraphics[width = 0.7\textwidth]{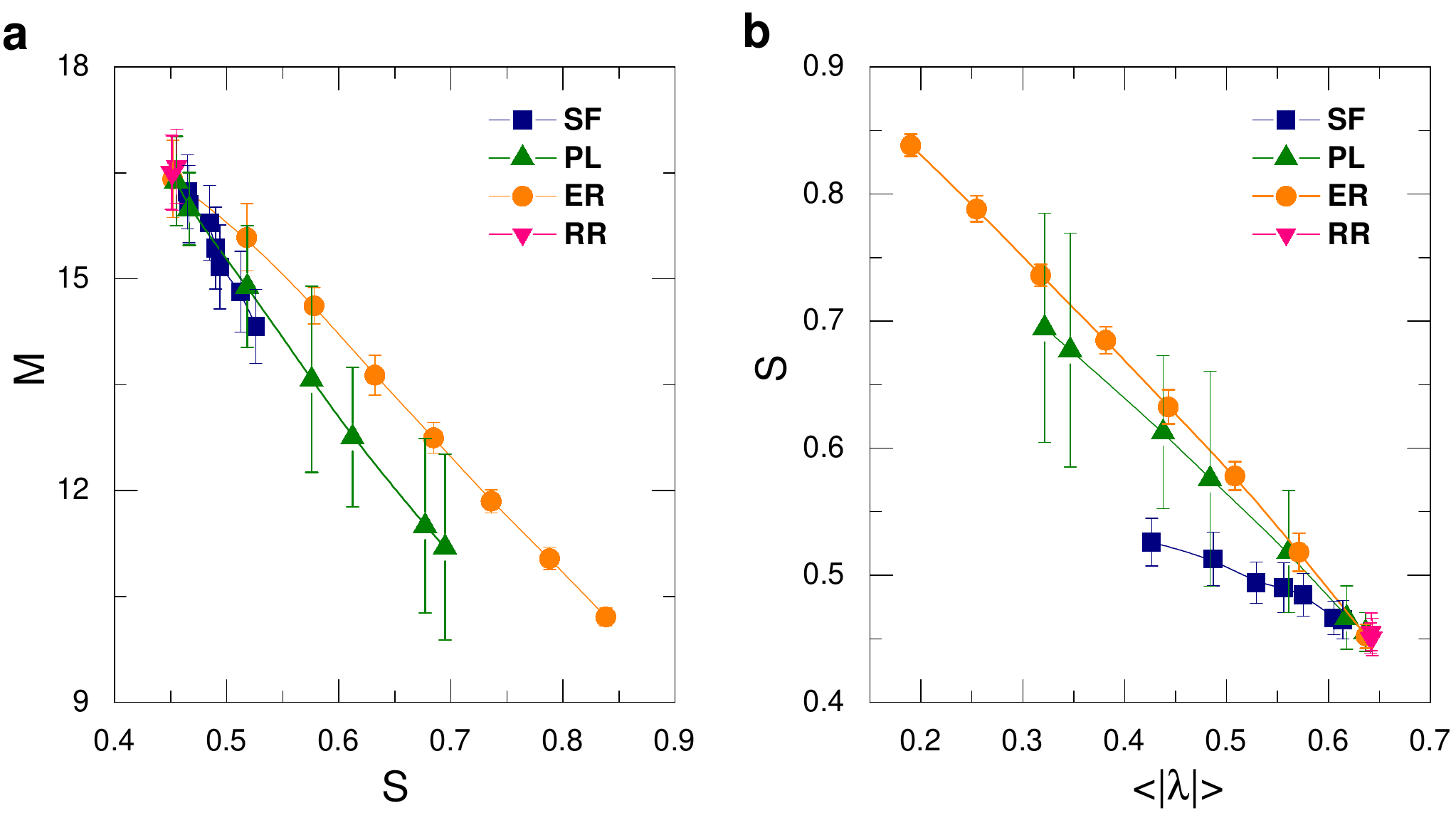}\\
	\caption{\noindent \textbf{Relationship between memory capacity, neuron correlation and the network spectrum}.(a) Memory capacity $M$ vs neuron state correlation $S$. (b) Neuron state correlation $S$ vs average eigenvalue modulus $\langle|\lambda|\rangle$.
	The ESNs were created using reservoirs of 400 neurons with spectral radius of $1$ and sequences of 4000 random inputs uniformly distributed in the interval $[-1,1]$. The ER curve is given by classical reservoirs having networks given by Erd{\"o}s-R{\'e}ny random graphs with weights drawn from a Gaussian distribution and varying spectral radii. The SF curve (blue) corresponds to scale-free networks where the degree heterogeneity is given by the degree exponent $\gamma\in [2, 6]$, with more heterogeneous networks rendering lower $M$, higher $S$ and lower $\langle\lambda\rangle$. The PL curve (green) is calculated from Erd{\"o}s-R{\'e}ny random graphs with weights drawn from a power-law (PL) distribution with varying exponent $\beta\in[2,5]$, with lower $\beta$ rendering lower $M$, higher $S$ and lower $\langle\lambda\rangle$. A more detailed numerical exploration of the dependency between the various network parameters and $M$, and between the network parameters and $\langle|\lambda|\rangle$ is presented in the Supplementary Information \ref{SI:numericsMemory}. All networks have a spectral radius $\alpha = 1$, except the ER random graphs where each point corresponds to a spectral radius in the range $\left[0.2,1\right]$ to show the impact of spectral radius. It is worth noticing that although the theoretical upper bound for the memory capacity of a reservoir is $M=N$\cite{jaeger2001short} and small input scalings \cite{farkavs2016computational} do achieve similar values, in our case input scaling is large and thus the nonlinearity of the reservoir limits $M$ to be less than 18.
		\label{fig:MemoryCorrLambda}}
\end{figure*}

\subsection{Memory and Structure in ESN}

The success of ESN in tasks such as forecasting time-series comes from the ability of its reservoir to retain memory of previous inputs\cite{cui2012architecture}. In ESN literature, this is quantified by the memory capacity defined as follows\cite{jaeger2001short}:
\begin{equation}\label{eq:Memory}
\begin{aligned}
M &= \sum_{\tau=1}^{N} M_{\tau},
\end{aligned}
\end{equation}
with $	M_{\tau} = \text{max}_{\mathbf{w}_{\text{out}}^\tau}\dfrac{\text{cov}^2(r(t-\tau),y_\tau(t))}{\text{var}(r(t-\tau)) \text{var}(y_\tau(t))}$. Here $r(t)$ is a random variable drawn from a standard normal distribution $\mathcal{N}(0,1)$, serving as a random input, `$\text{cov}$' represents the covariance, $y_\tau(t)$ is the output as described in Eq. \ref{eq:ESNReadout}, $\mathbf{w}_{\text{out}}^\tau$ is obtained as a minimizer of the difference between $y_\tau(t)$ and $r(t-\tau)$.

To quantify the relationship between the reservoir dynamics and the memory capacity, we note that the extraction of information from the reservoir is made through a linear combination of the neurons' states. Hence it is reasonable to assume that more linearly independent neurons would offer more variable states, and thus longer memory\cite{lukovsevivcius2009reservoir, jaeger2005reservoir}. In plain words, we hypothesize that the memory capacity $M$ strongly depends on the correlations among neuron states, which can be quantified as follows:
\begin{equation}\label{eq:CorrelationCoeff}
\begin{aligned}
S &= \dfrac{\sum_{i=1}^{N-1}\sum_{j=i+1}^{N}P_{ij}^2}{N(N-1)/2}.
\end{aligned}
\end{equation}
Here $P_{ij} = \dfrac{\text{cov}\left(x_i(t),x_j(t)\right)}{ \text{std}(x_i(t))\text{std}(x_j(t))}$ is the Pearson's correlation coefficient between the states of neurons $i$ and $j$, and $\text{std}(x_i)$ represents the standard deviation. Thus, $S$ is simply the average of those squared correlation coefficients, representing a global indicator of the correlations among neurons in the reservoir. Fig. \ref{fig:MemoryCorrLambda}a corroborates our hypothesis, showing that for various network topologies there is a strong correlation between $S$ and $M$, which can also be justified analytically (see Supplementary Information \ref{SI:analyticMemory}). Thus, hereafter we only need to understand how the network structure affects the neuron correlation.

\begin{figure*}[ht!]
	\centering
	\includegraphics[width=0.8\textwidth]{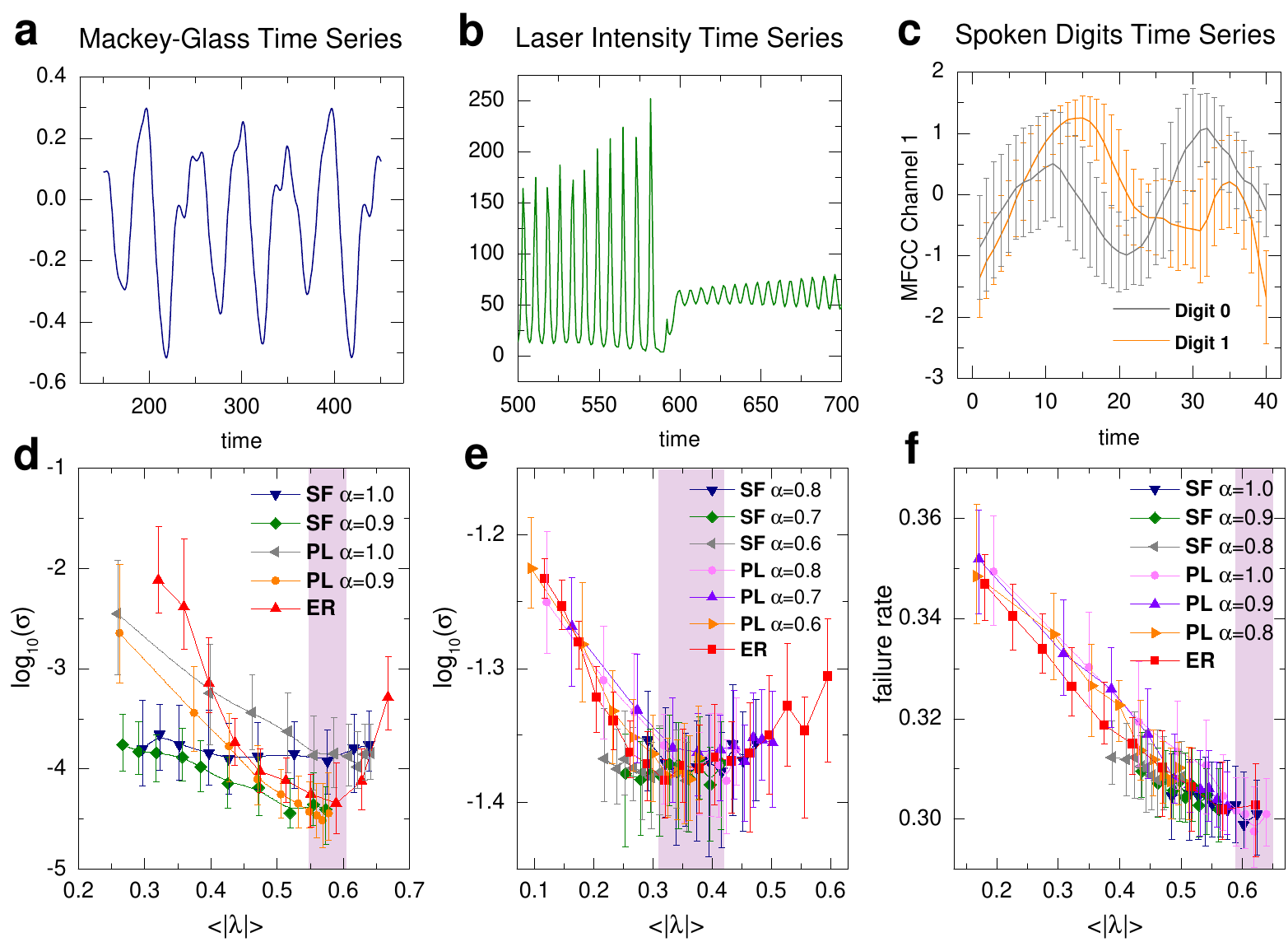}\label{fig:perfVsLambda}\\
	\caption{\noindent \textbf{Time series analyzed in this work and the ESN performance explained by $\langle|\lambda|\rangle$}. The plot shows samples of the three datasets used (a-c) and their associated ESN performance as a function of the average eigenvalue modulus $\langle|\lambda|\rangle$ (d-f). (a) the Mackey-Glass time series with 500 points. (b) the Laser Intensity time series with 300 points. (c) the average value of the first mel-frequency cepstral coefficient (MFCC) Channel of the first Spoken Arabic Digit, the error bars represent standard deviations over the training dataset. (d) the ESN forecasting performance for the Mackey-Glass time series, (e) the forecasting performance for Laser Intensity time series, and (f) the failure classification rate for Spoken Arabic Digits. For each task, we use scale-free networks (SF), Erd\H{o}s-R\'enyi random graphs with homogeneous link weights (ER), and Erd\H{o}s-R\'enyi random graphs whose link weights follow a power-law distribution (PL) as reservoirs  (see SI Sec. II for the network generation algorithms). The SF and PL reservoirs have various spectral radii $\alpha$, chosen to be around the optimal value of $\alpha$ for the Erd\H{o}s-R\'enyi case. For each parameter set of each network type we created 200 ESN realizations, and then all the points obtained were grouped in 10 bins containing the same number of points. We plotted their median $\langle|\lambda|\rangle$ against their median performance: $\sigma$ from Eq.\ref{eq:NRMSE} for (d) and (e); and the failure rate for (f), with the error bars being the upper and lower quartile respectively. The three panels show that, regardless of topology or spectral radius, all networks have their optimal performance when the average eigenvalue module, $\langle|\lambda|\rangle$, is within the intervals $[0.55, 0.6]$ (d), $[0.3, 0.4]$ (e), or $[0.6, 0.7]$ (f), which are highlighted in pink. See the Appendix \ref{app:PerfMeasurement} for an expanded description of the three datasets and the performance measurement.
		\label{fig:SpectrumAndPerformance}}
\end{figure*}

We consider the neuron correlation as a measure of coordination. If the neuron states are highly interdependent, then $S$ will be very high. By contrast, a system with independent neurons will have very low $S$. For linear dynamical systems, the correlations between neurons depend on all the eigenvalues of the adjacency matrix\cite{boccaletti2006complex}, with larger mean eigenvalue meaning lower correlations (see Supplementary Information \ref{SI:analyticMemory}). Our system (\ref{eq:ESNModel}) is non-linear, but its first order approximation is the identity function $f(z)=z$. Hence we can use the eigenvalues of matrix $\mathbf{W}$ to approximately quantify how fast the input decays in the reservoir, and hence how poorly the ESN remembers. 

To quantify the aggregated effect of the eigenvalue distribution in the complex plane, we define the average eigenvalue moduli:
\begin{equation}\label{eq:lambda}
\begin{aligned}
\langle|\lambda|\rangle = \dfrac{\sum_{i=1}^{N}|\lambda_i|}{N},
\end{aligned}
\end{equation}
where $\lambda_i$'s are the eigenvalues of the matrix $\mathbf{W}$. As expected from our previous discussion, we find that $\langle|\lambda|\rangle$ strongly correlates with $S$ (Fig. \ref{fig:MemoryCorrLambda}b). The correlation of $\langle|\lambda|\rangle$ with $S$ and therefore with $M$ indicate that $\langle|\lambda|\rangle$ indeed reflects the memory capacity of the reservoir. As opposed to $M$ and $S$, $\langle|\lambda|\rangle$ is much easier to compute and is solely determined by the reservoir network. This offers a simple measure to quantify the ESN memory capacity that does only depend on the network structure. $\langle|\lambda|\rangle$ is consistent with the effects of scaling the adjacency matrix to tune the spectral radius\cite{jaeger2007optimization} and it extends to network topologies with a fixed spectral radius (see Supplementary Figures 2, 3). This also explains two recent studies in which it was found that ring networks and orthogonalized networks have high memory capacities \cite{rodan2012simple,farkavs2016computational}, as both networks have large eigenvalues with respect to their spectral radii. Finally, it suggest that $M$ can be maximized by using networks with large eigenvalues, the simplest one being a circulant network with degree $1$ \cite{aceituno2018eigenvalues}, which achieves $M=20$ while the others go only up to $M=17$ (see the example in Supplementary Information \ref{SI:MaximizeM}).

As ESN is fundamentally a machine learning tool, any results should be supported by studying its performance at different tasks. Here we chose the following three tasks: (1) forecasting the chaotic Mackey-Glass time series \cite{mackey1977oscillation}, which is a classical task in ESN \cite{jaeger2004harnessing, lukovsevicius2007overview}, (2) forecasting the Laser Intensity Time series \cite{hubner1989dimensions} downloaded from the Santa Fe Institute; and (3) classifying Spoken Arabic Digits \cite{hammami2010improved} downloaded from the Machine Learning Repository of the UCI \cite{Lichman:2013} (see Figure 3a-c). To demonstrate the validity of $\langle |\lambda| \rangle$ as a proxy measure for memory, we tested ESN performance for the three tasks with a wide range of network topologies and parameters. As each task requires a memory capacity that is independent of the type of network, we find that the optimal parameters for all networks are within a consistent range of $\langle |\lambda| \rangle$ (see Figure \ref{fig:SpectrumAndPerformance}d-f).

\subsection{Adapting ESN in the frequency domain}
	\begin{figure*}[ht!]
	\centering
	\includegraphics[width=1\linewidth,trim = {0cm 0cm 0cm 0cm}, clip]{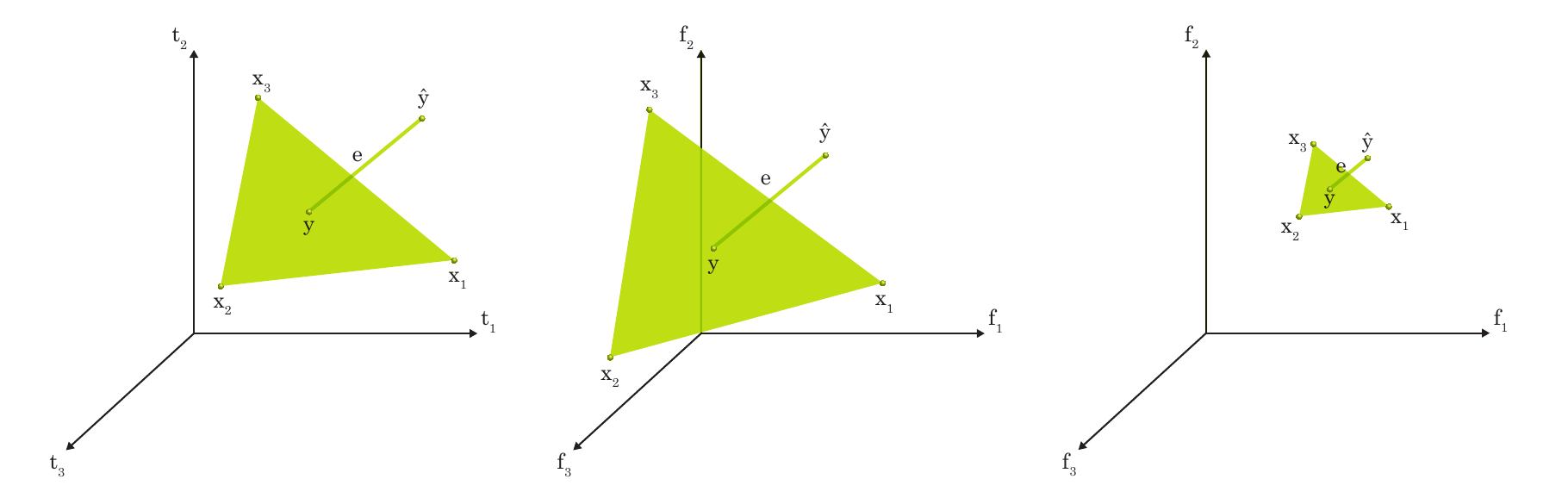}\\
	\caption[Sketch of the frequency adaptation argument]{\noindent \textbf{Sketch of the frequency adaptation argument}: The target output of the training is a time series of length $T$, which can be expressed as a vector in the corresponding space. A reservoir consists of $N$ non-linear filters of the input time series, which can generate $N$ points in that same space (left); the readout simply selects the point in the subspace spanned by the $N$ neurons that is closest to the target. Since the distances between vectors do not change after applying the Fourier Transform -- by Parseval's Theorem \cite{parseval1806memoire} --, the picture is still valid in the Fourier Domain (center). However, in that domain it will be possible to alter the filters so that the $N$ points approach the target by making the reservoir resonate at the appropriate frequencies, effectively reducing the error (right).}
	\label{fig:freqAdaptationArgument}
\end{figure*}

Intuitively, a reservoir can be understood as a set of coupled filters that extract features from the input signal, and the readout simply selects the right combination of those features. In the remaining of this section we will show that the filters should be designed to extract the features that are relevant for the problem at hand, and that these features can be expressed in the Fourier domain. 

This can be translated to machine learning terms through a geometric argument presented in Fig.~\ref{fig:freqAdaptationArgument}, and is mare rigorous in Supplementary Information \ref{SI:whyFreq}, where we derive the bound
\begin{equation}\label{eq:FrequencyBound}
\begin{aligned}
\sigma &\leq \dfrac{\sum_{n=1}^{N} | \mathcal{F}[x_n]\times \mathcal{F}[\hat{y}]|}
{\sum_{n=1}^{N}| \langle\mathcal{F}[x_n],\mathcal{F}[\hat{y}]\rangle|}
\end{aligned}
\end{equation} 
where $\mathcal{F}\left[\cdot\right]$ stands for the Fourier Transform, $\times$ and $\langle \cdot,\cdot\rangle$ are the cross and scalar product, and $\sigma$ is the Normalized Root Mean Square Error (NRMSE) presented in Appendix~\ref{app:PerfMeasurement}. 

In terms of signal processing, $ | \mathcal{F}[x_n]\times \mathcal{F}[\hat{y}]|$ and $|\langle\mathcal{F}[x_n],\mathcal{F}[\hat{y}]\rangle|$ can be expressed by how much the power spectral densities (PSD) of the neurons resemble -- for the scalar product -- or differ --for the cross product -- from $y$. This is quite a natural result, as it simply implies that if the time series of the variables $\mathbf{x}$ are similar to the target, then the readout will work better. 

\begin{figure*}[ht!]
	\centering
	\includegraphics[width=1\textwidth,height=0.7\textwidth]{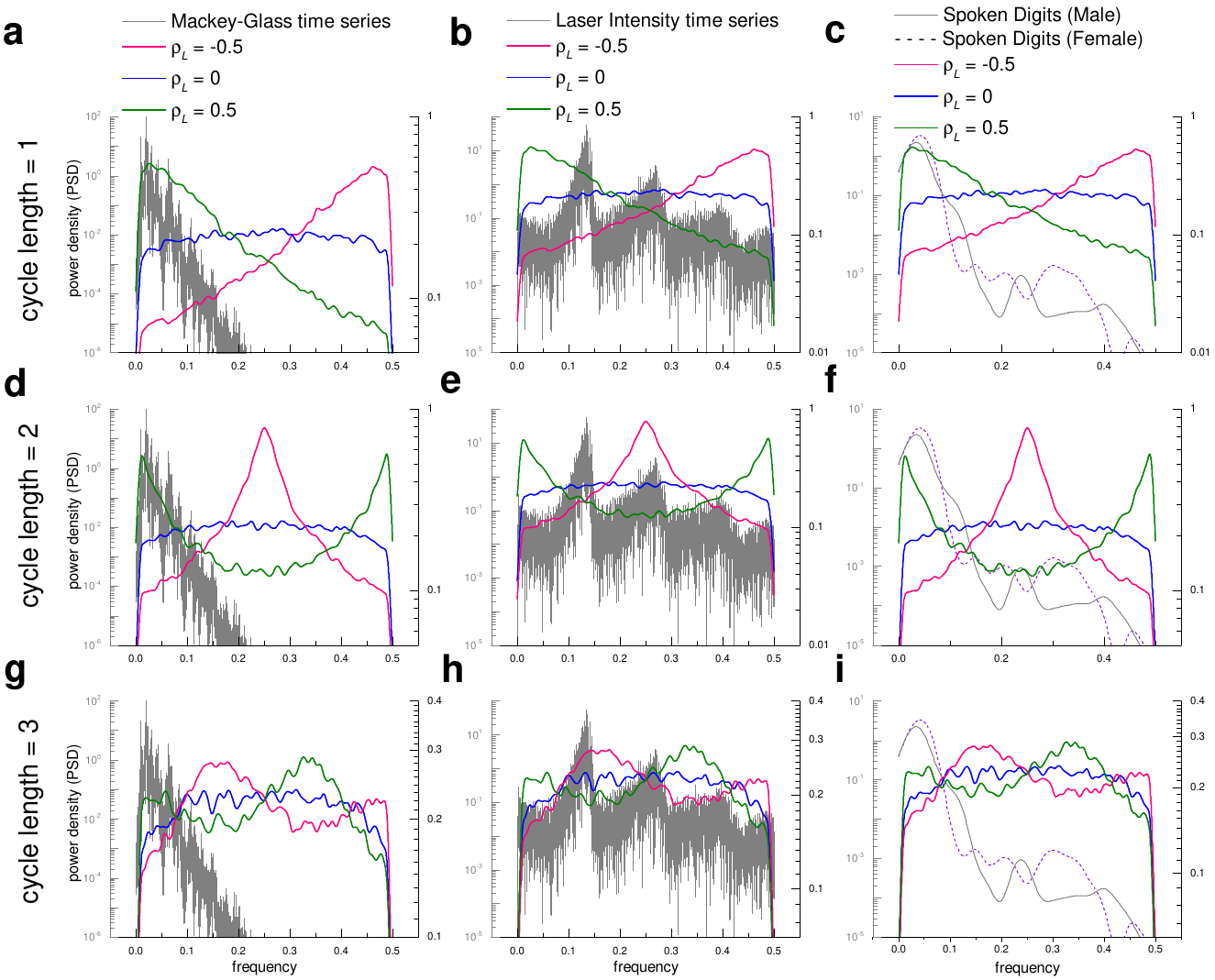}\\
	\caption{\noindent \textbf{Frequency domain analysis of target signals and reservoir frequencies}. We plotted the power spectral density (PSD) of three empirical time series (Mackey-Glass in a,d,g; Laser Intensity in b,e,h; Spoken Digits in c,f,i), and the average PSD of the reservoirs' neuron states for reservoirs with various $\rho_L$ when using a random Gaussian input with zero mean and variance of one (left y-axis for PSD of empirical time series and right y-axis for PSD of reservoirs with random inputs). In each panel we plot the average PSD of 500 reservoirs with 400 neurons and connectivity $0.05$.	The length of cycles added into the reservoir is 1 (a-c), 2 (d-f) and 3 (g-i).
	Note that changing the reservoir through adding cycles can make the response of neurons similar to the signals'. In the case of Mackey-Glass and Spoken Arabic Digit, the signals are defined mostly by frequencies close to 0. Any reservoir with $\rho_L>0$ will enhance the frequencies close to 0, meaning that such a reservoir would enhance the frequencies relevant to those two time series. Similarly, the Laser Intensity has three peaks that are closer to the center of the spectrum, which are enhanced in the cases of $\rho<0$ for $L=2$ and $L=3$, but not for $L=1$, in which case neither $\rho>0$ nor $\rho<0$ will enhance the three peaks. This will have affect the performance of ESN (see Fig.~\ref{fig:cyclesPerf}).}
	\label{fig:freqEnhancement}
\end{figure*}

Thus we need to alter the PSD of the reservoir, which we do by adding feedback loops with delay $L$ in our neurons, encoded as cycles of length $L$ in the network. We account for the strength of those cycles by using the following measure:
\begin{equation}\label{eq:rho}
\begin{aligned}
\rho_L = \frac{E_{L,s} - E_{L,-s}}{E},
\end{aligned}
\end{equation}
where $E_{L,s}$ is the number of edges embedded in cycles of length $L$ and sign $s$ which is fed into the network generation algorithm presented in the Supplementary Information \ref{SI:netWithCycles}. We show numerically that a reservoir with those cycles enhances different families of frequencies to reservoirs with different $\rho_L$ values, as shown in Fig.~\ref{fig:freqEnhancement}. This works even though the reservoir is a nonlinear system, by the monotonicity of the nonlinearity, as we analyze in Supplementary Information \ref{SI:analyticFreq}.

Knowing that altering the frequencies of the reservoir can improve its performance and that adding cycles can adapt those frequencies, the next natural step is to apply this knowledge and design reservoirs adapted to a specific task. We start by simply taking different reservoirs with varying fractions of cycles, and compare values of $\rho_L$ against the performance of the reservoir in that task.

\begin{figure*}[ht!]
	\centering
	\includegraphics[width=1\textwidth]{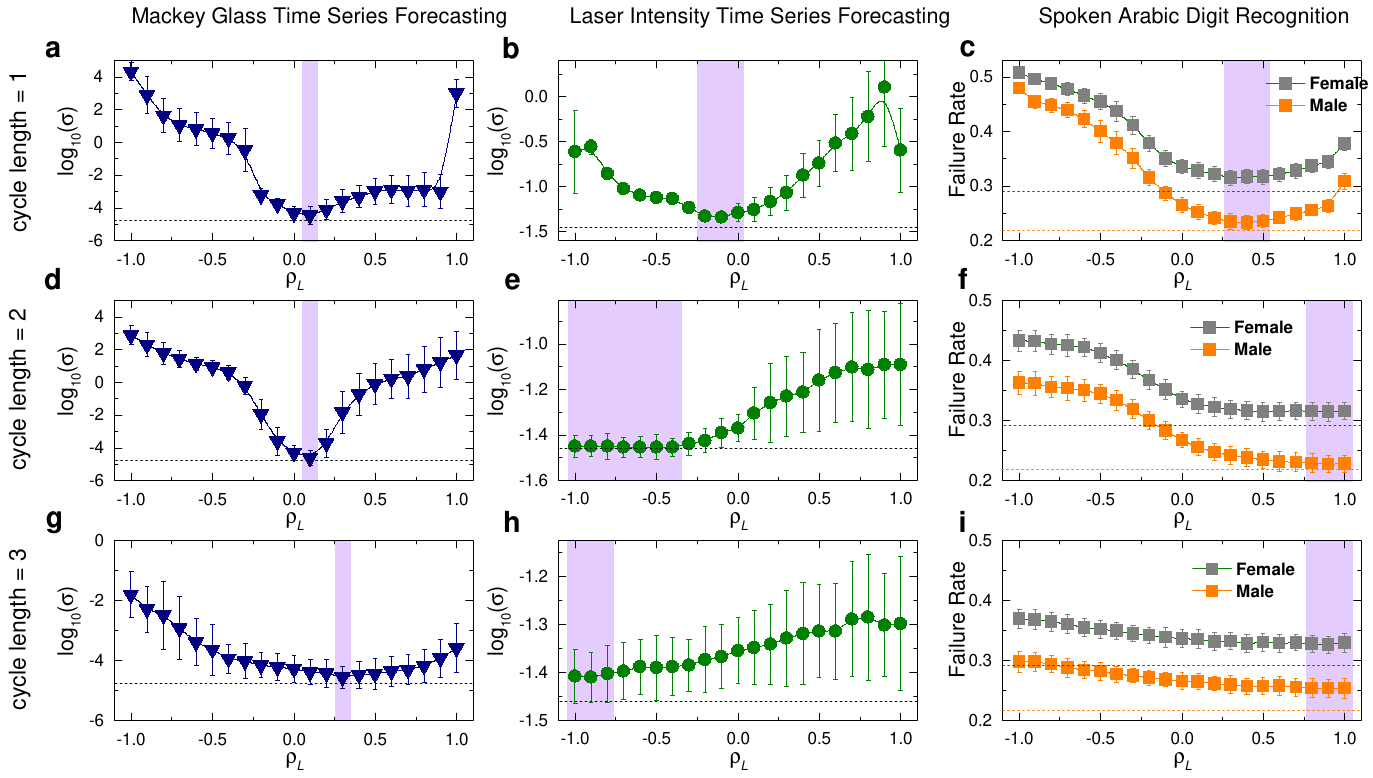}\\
	\caption[Improving ESN through frequency adaptation]{\noindent \textbf{Improving ESN through frequency adaptation}. ESN performance $\sigma$ vs $\rho_L$, for the tasks of Mackey-Glass Forecasting (a, d, g), Laser Intensity Forecasting (b, e, h), and Spoken Arabic Digit Recognition (c, f ,i). The length of cycles added into the reservoir is 1 in (a-c), 2 in (d-f) and 3 in (g-i). Every point corresponds to the median performance --measured by NRMSE in the (a, b, d, e, g, h) and by the failure rate in (c, f, i)-- over 200 ESN realizations with the error bars corresponding to upper and lower quartiles. Following our observation in Fig.~\ref{fig:freqEnhancement}, the performance of ESN increases when the the reservoir enhances the frequencies that are dominant in the signal. Thus, since the Mackey-Glass and Spoken Arabic Digit signals have low frequencies, the performance of ESN for those signals increases when $\rho>0$. In contrast, the Laser Intensity time series is dominated by frequencies that are on the center of the spectrum, thus ESN has a better performance when $\rho_1\simeq 0$, $\rho_2<0$ and $\rho_3<0$, as those parameters adapt the reservoir's response to the time-series frequencies.
	In all cases we see that combining cycles of different lengths can improve ESN performance.
		\label{fig:cyclesPerf}}
\end{figure*}

As we see in Fig. \ref{fig:cyclesPerf}, the performance of ESN changes with the fraction of cycles $\rho_L$. To understand them better, it is useful to compare the optimal $\rho_L$ values with the PSD of the time series. A simple example is given by the Mackey-Glass time series: Fig. \ref{fig:cyclesPerf} shows that for $\rho_L>0$, the reservoir's average PSD response is enhanced for the frequencies close to $0$, which is exactly the regime where the spectrum of the Mackey-Glass Time Series is concentrated \ref{fig:freqEnhancement}. Similarly, the Spoken Arabic Digits are also dominated by frequencies close to 0, and thus the performance of ESN improves when $\rho_L>0$ (Fig. \ref{fig:cyclesPerf}c,f,i). As for the Laser Intensity Time Series, its dominating frequencies are around $0.13$, $0.27$ and $0.38$, thus ESN is improved when the response of the reservoir enhances those frequencies. As shown in Fig.~\ref{fig:freqEnhancement}.e,h, this happens when $\rho_L<0$ for $L=2,3$. Indeed, as shown in Fig.~ \ref{fig:cyclesPerf}e,h, negative $\rho_L$ (for $L=2,3$) improves the ESN performance. For the case of $L=1$, we observed in Fig. \ref{fig:freqEnhancement} that the three peaks cannot be all enhanced simultaneously by setting $r_1$ to be either positive or negative. Instead, setting $r_1 = 0$ would yield the optimal performance. This is what we observe in Fig.~\ref{fig:cyclesPerf}.b.

The immediate conclusion is that the reservoir should be designed to enhance the frequencies present in the target signal. A simple way of achieving this is to obtain the frequency responses of the reservoir, and then select the parameters of the reservoir for which the frequencies match the target signal better. Based on those considerations we designed a simple heuristic algorithm to find the optimal values of $\rho_L$ for cycles of different lengths, which we describe in the Supplementary Information~\ref{SI:designFreq}. This heuristic does indeed improve the performance of ESN beyond that of random reservoirs, as shown in Fig.~\ref{fig:cyclesPerf} in the dotted lines.

\section{Discussion}
In this paper we explore how simple ideas from classical signal processing and network science can be applied to dissect a classical type of recurrent neural network, i.e., the ESN. Moreover, dissecting ESN helps us design simple strategies to further improve its performance by adapting the dynamics of its reservoir. We find that the memory capacity depends the full spectrum of the connectivity matrix $\mathbf{W}$, generalizing the spectral radius as a memory parameter. Moreover, we demonstrate that the PSD of the time series is very important, and we provide simple tools to adapts the reservoir to a specific frequency band.

It is important to note that we are not advocating for hand-tunning reservoir topologies for specific tasks, but rather to raise the point that notions from classical signal processing can help us understand and improve recurrent neural networks, either through selection of appropriate initial topologies in a pre-training stage, or by designing learning algorithms that account for the principles outlined here. Given that most current learning strategies such as backpropagation focus on adapting single weights, we are convinced that many new learning algorithms can be created by focusing on network-level features. However, our approach goes beyond improving current techniques. By studying which properties of a recurrent neural network make it well-suited for a particular problem, we are also addressing the converse question of how should a neural network be after it has been adapted to a specific task. Thus, we provide valuable insights into the training process of general recurrent neural networks, as our theory highlights structural features that the training process would enhance or inhibit.

To conclude, our results presented here clearly indicate that notions from physics, dynamical systems, control theory and network science can and should be used to understand the mechanism and then improve the performance of artificial neural networks. 

\vspace{2cm}

\appendix

\section{Performance Measurement}\label{app:PerfMeasurement}

In the literature of ESN it is common to forecast time series \cite{jaeger2004harnessing}. To be consistent with the previous literature we use the normalized root mean squared error (NRMSE), as a metric of forecasting error 
\begin{equation}\label{eq:NRMSE}
\sigma = \sqrt{\dfrac{\sum_{t=t_0}^{t_0+T} (y(t)-\hat{y}(t))^2}{T\cdot \textrm{var}(\hat{y}(t))}}.
\end{equation}
This metric is a normalization of the classical root mean squared error. The parameter $t_0$ is used to describe when we start to count the performance, since it is also common to ignore the inputs during the initialization phase \cite{jaeger2002tutorial}, which is taken here as the full initialization steps given for each task (see details in the subsequent sections). The parameter $T$ is simply the number of time-steps considered, which we take here as the full count of all points except the initialization phase in each testing time series. 

The NRMSE is obviously not a good metric for classification tasks where the target variable is discrete. In order to have a comparable metric for ESN performance, we use the failure rate in classification tasks such as the Spoken Arabic Digit Recognition. Note that having 10 digits implies that the failure rate with random guesses is 0.9, therefore a failure rate of 0.3 is well below it.

\subsection{Forecasting Mackey-Glass time series}
Forecasting Mackey-Glass time series is a benchmark task to test the performance of ESN \cite{jaeger2004harnessing}. The Mackey-Glass time series follows the ordinary differential equation\cite{jaeger2004harnessing}: 
\begin{equation*}\label{eq:MGEq}
\begin{aligned}
\frac{ds(t)}{dt} = \beta \frac{ s(t-\tau) }{1+{s(t-\tau)}^n}-\gamma s(t), 
\end{aligned}
\end{equation*}
where $\beta$, $\gamma$, $\tau$, $n$ are real positive numbers. We used the parameters $\beta=0.2$, $\gamma=0.1$, $\tau=17$, $n=10$ in our simulations. The discrete version of the equation uses a time step of length $h=0.1$. For each time series we generated $\frac{\tau}{h}=170$ uniformly distributed random values between $1.1$ and $1.3$ and then followed the equations. The first 1000 points were considered as initialization steps, which did not fully capture the time series dynamics and were thus discarded. 
For training and testing we used time series of 10,000 points, but in both cases the first 1000 states of the reservoir were considered as initialization steps and were thus ignored for training and testing.
For an ESN with 1000 neurons and an optimized memory, the forecasting performance for this setting is close to its maximum value, thus the addition of short cycles will have a small effect. In order to show the interest of our contribution, we normalized the signal to have mean zero and variance of one and we added Gaussian white noise with $\sigma = 0.05$, and the forecasting was done using reservoirs of 100 neurons, average degree $\langle k\rangle=10$ and spectral radius of $\alpha=0.85$, and the output was feed back to the reservoir through the vector $\mathbf{w}^{\text{ofb}}$ where every entry is independently drawn from a uniform distribution on the interval $[-1,1]$. The ESN was trained to forecast one time-step, and then we used this readout to forecast 84 time-steps in the future by recursively feeding the one-step prediction of $s(t+1)$ into the ESN as the new input. 

\subsection{Forecasting Laser Intensity time series}
The Laser Intensity time series \cite{hubner1989dimensions,huebner1989problems} was obtained from the Santa Fe Institute time series Forecasting Competition Data. It consists of 10,093 points, which we normalized to have an average of zero and an standard deviation of one, and were filtered with a Gaussian filter of length three and standard deviation of one.
The forecasting was done using reservoirs of 100 neurons, average degree $\langle k\rangle=10$ and spectral radius of $\alpha=0.9$, without feedback so $\mathbf{w}^{\text{ofb}} = 0$. Here we forecasted one time-step.
We used 1,000 points of the time series for initialization, 4,547 for training and 4,546 for testing. 

\subsection{Spoken Arabic Digit Recognition}
The Spoken Arabic Digits \cite{hammami2010improved} dataset was downloaded from the \cite{hammami2010improved} from the UCI Machine Learning Repository \cite{Lichman:2013}. This dataset consists of 660 recordings (330 from men and 330 from women) for each of the ten digits and 110 recordings for testing. Each recording is a time series of varying length encoded with MCCF \cite{mermelstein1976distance} with 13 channels. While using the first three channels gave a better performance, here we use only the first channel, which is akin to a very lossy compression. We normalized this time series to have average of zero and a standard deviation of one, and a length of 40. Since in most cases we had less than 40 points, we computed the missing values by interpolation.
The classification procedure was done using the forecasting framework. We collected the reservoir states from all the training examples of each digit and computed $\mathbf{w}^{\text{out}}$ as did in the previous forecasting tasks. In the testing we collected the states and computed the forecasting performance $\sigma$ for each of the 20 cases. We classified the time series as the digit that yielded the lowest forecasting error. Then we calculate the failure rate as the number of misclassified recordings divided by the total number of recordings in the training set.
We used reservoirs of 100 neurons, average degree $\langle k\rangle=10$ and spectral radius of $\alpha=1$, without feedback ($\mathbf{w}^{\text{ofb}} = 0$).

\section{Numerical study of eigenvalue density and memory capacity}\label{SI:numericsMemory}
\begin{figure*}
	\centering
	\includegraphics[width=1\linewidth,trim={2.5cm 0 3cm 0},clip]{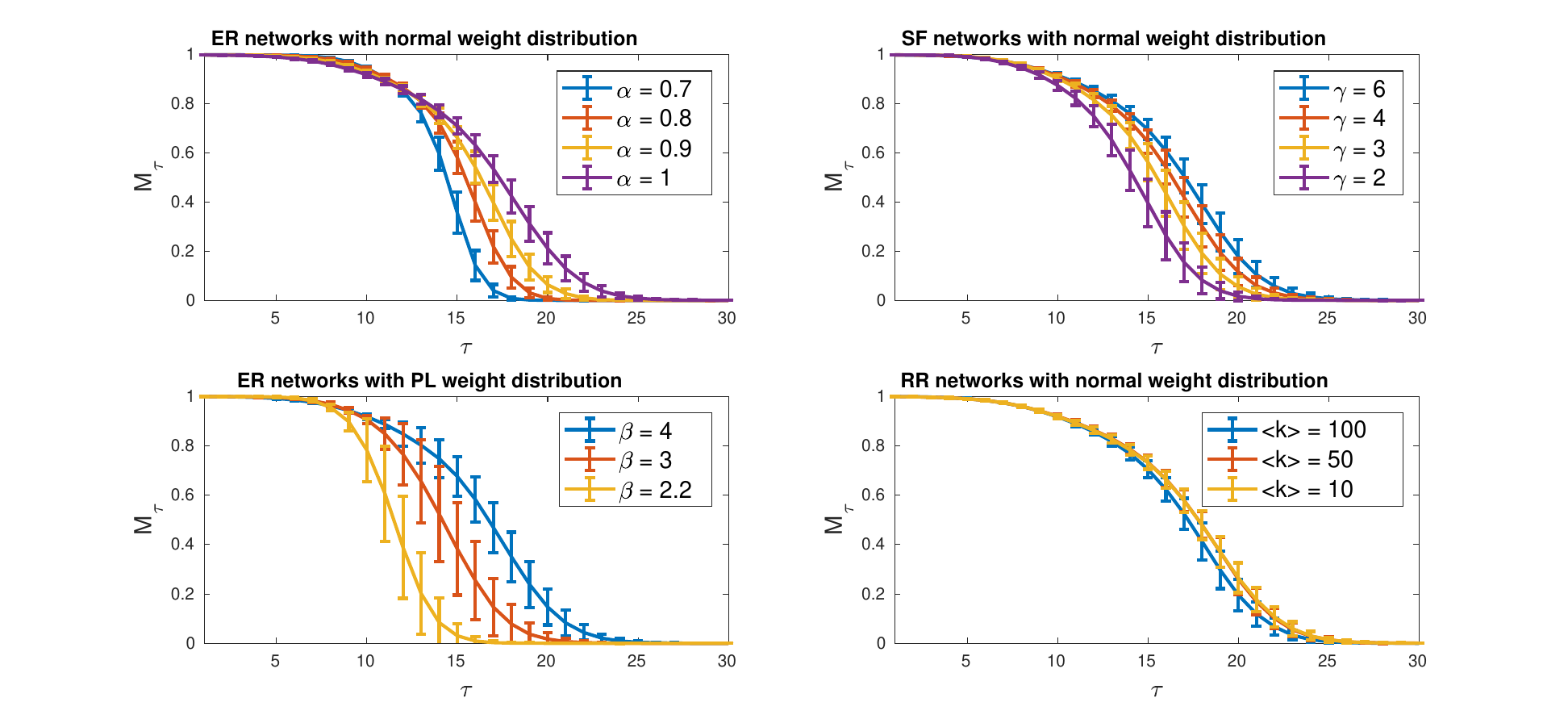}
	\caption{The ability of the reservoir to retrieve previous inputs decays as with $\tau$, the delay with which we try to retrieve them. Each network has 400 neurons and average degree is $\langle k \rangle = 20$. Each curve shows the average result over 100 trials and the error bars are the standard deviation. The spectral radius $\alpha = 1$, except in ER networks where it varies as marked in the legend.}
	\label{fig:MemoryCurves}
\end{figure*}

One of the motivations of our study on Memory Capacity is that the spectral radius is not enough to capture how $M$ changes, particularly for different network structures. In Fig.\ref{fig:MemoryCurves}, we show that for some families of networks the memory is also affected by other parameters. 

Specifically, we study the following architectures:
\begin{itemize}
	\item Erd{\"o}s-R{\'e}ny (ER) random graphs with weights drawn from a Gaussian distribution and varying spectral radii.
	\item Erd{\"o}s-R{\'e}ny random graphs with weights drawn from a power law distribution (PL) with $\beta\in[2,5]$ but normalized to have a spectral radius $\alpha = 1$.
	\item Scale-Free (SF) networks where the degree heterogeneity is given by the degree exponent $\gamma\in [2, 6]$, and the weights are drawn from a Gaussian distribution, also normalized to have a spectral radius $\alpha = 1$.
	\item Random Regular (RR) graphs with varying degrees and a spectral radius $\alpha = 1$.
\end{itemize}

\begin{figure*}
	\centering
	\includegraphics[width=1\textwidth,trim={2cm 0 2cm 0},clip]{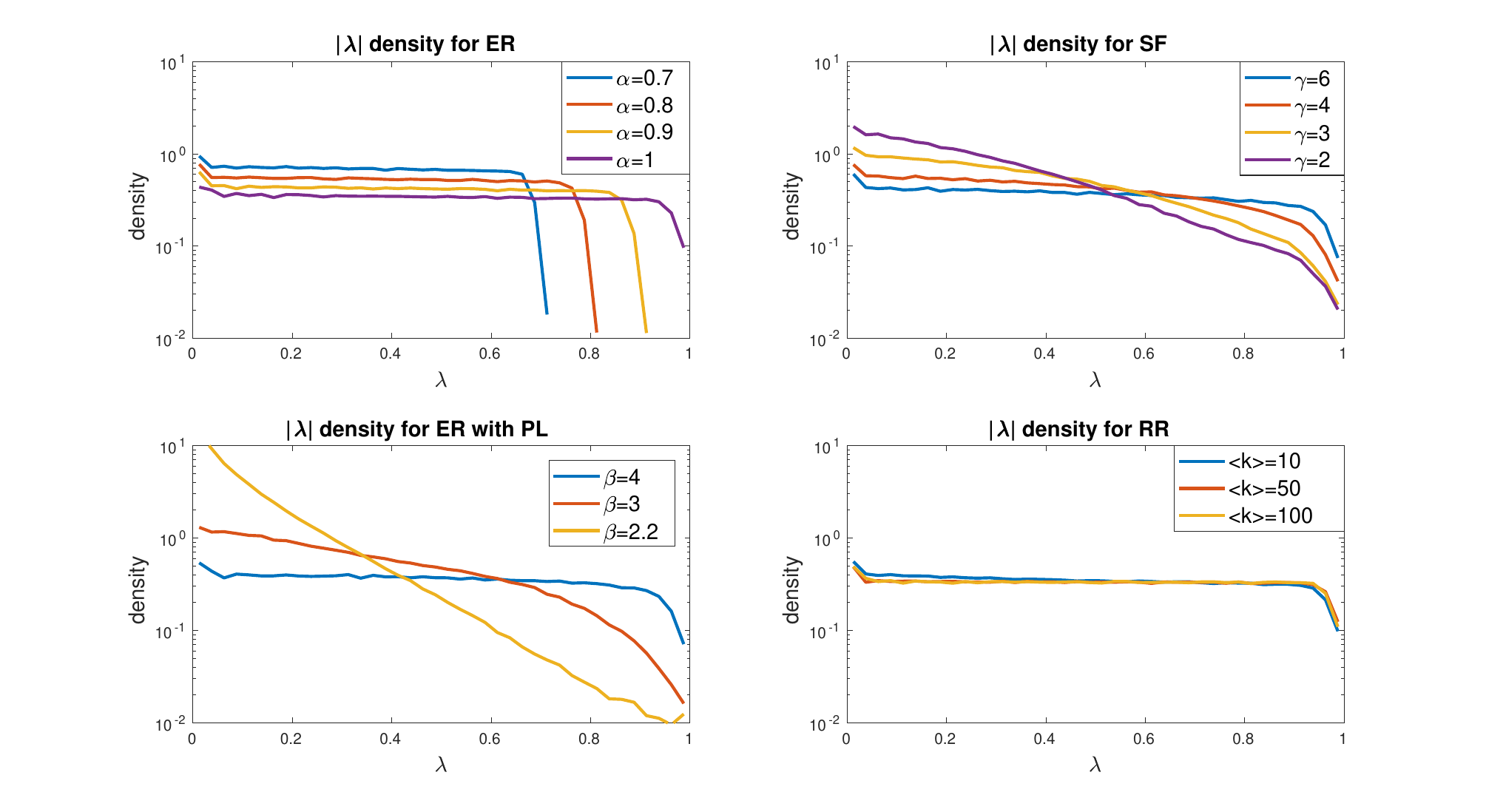}
	\caption{ Eigenvalue densities of the random networks studied in the main text. Each line shows the density of eigenvalues at a distance $\lambda$ from the origin. $200$ realizations of a network. The network size is $1000$, the edge weights are drawn from a normal distribution except when in the ER networks with PL. Unless explicitly stated in the sub-figure legend, $\langle k\rangle=50$ and the spectral radius is $\alpha = 1$.}
	\label{fig:NetworkSpectra}
\end{figure*}

The results from Fig.\ref{fig:MemoryCurves} can be contrasted with the eigenvalue densities of the aforementioned network families presented in Fig.\ref{fig:NetworkSpectra}. We observe that the networks where the eigenvalues are concentrated in the center, either through the spectral radius $\alpha$, the power law exponent $\beta$ or the degree heterogeneity $\gamma$, have lower memory, while those with eigenvalues uniformly spread have more memory.

\section{Memory Capacity and the Correlations Between Neurons}\label{SI:analyticMemory}

Here we derive an upper bound which connects the variance of the reservoir across different directions with the memory capacity. The gist of our argument goes as follows: the memory capacity reflects the precision with which previous inputs can be recovered; the nonlinearity of the reservoir and other, far-in-the-past inputs induce noise that complicates recovery, so the variance of the linear part of the reservoir must be placed in such a way as to maximize how much information can be recovered. First we argue that the inputs should be projected into orthogonal directions of the reservoir state space, so that they do not add noise to each other; within those orthogonal directions, the variance should be such that it is as evenly spread across the different dimensions --inputs -- as possible, as concentrating the variance into few inputs makes the rest loose precision. This is quantified by the covariance of the neurons.

We start by noticing that the linear nature of the projection vector $\mathbf{w}_{\text{out}}$ implies that we are treating the system as 
\begin{equation}\label{eq:linearViewOnRC}
\mathbf{x}(t) = \sum_{k=0}^{\infty}\mathbf{a}_kr(t-k) + \boldsymbol{\varepsilon}_r(t)
\end{equation}
where the vectors $\mathbf{a}_k\in \mathbb{R}^N$ are correspond to the linearly extractable effect of $r(t-k)$ onto $\mathbf{x}(t)$ and $\boldsymbol{\varepsilon}_r(t)$ is the nonlinear contribution of all the inputs onto the state of $\mathbf{x}(t)$. 

Notice that there are two perspectives here: on one side, the readout extracts the best linear approximation of past inputs with a noise-like term, and on the other it can be interpreted as a Taylor expansion around an undefined point where the first order corresponds to the first term and the non-linear behavior to the other expansion terms. This separation between linear and non-linear behavior has been thoroughly studied \cite{dambre2012information,ganguli2008memory} and the general understanding is that linear reservoirs have longer memory, but nonlinearity is needed to perform interesting computations. Here we will not try to leverage or bypass this trade-off, but rather we will show that for a fixed ratio of the non-linearity, the more decorrelated the neurons are the higher the memory.

To maintain this trade-off between linear and non-linear behavior, we will assume that the distribution of the linear and non-linear strengths are fixed. This can be achieved if we impose the probabilities of the neuron states do not change, meaning that the mean, variance and other moments of the neuron outputs are unchanged and hence the strength of the nonlinear effects is unchanged. 

A first constraint can also be obtained from the maintained strength of the linear side of Eq.~\ref{eq:linearViewOnRC}
\begin{equation}\label{eq:constraintMem2_naive}
\text{var}\left(\sum_{\tau=1}^{\infty} \mathbf{a}_\tau r(t-\tau)\right) = c
\end{equation}
where $c$ is a constant. 

If we are allowed to shift the linear side of Eq.~\ref{eq:linearViewOnRC}, the natural choice of $\mathbf{a}_\tau$ to maximize the memory capacity would be to impose $a_\tau>0\ \iff \tau>N$, and also make the vectors $\mathbf{a}_\tau$ orthogonal to each other, so that the input at time $t-\tau_1$ does not interfere with the effect of an input at time $t-\tau_2$. This can be introduced into Eq.~\ref{eq:constraintMem2_naive} and we obtain
\begin{equation}\label{eq:constraintMem1_1}
\text{var}\left(\sum_{\tau=1}^{\infty} \mathbf{a}_\tau r(t-\tau)\right) = 
\sum_{\tau=1}^{N} \text{var}(r(t-\tau))\|\mathbf{a}_\tau\|^2 
= \sum_{\tau=1}^{N} \|\mathbf{a}_\tau\|^2 = c
\end{equation}

This leaves us with a straightforward choice for the readout vector, namely
\begin{equation}
\mathbf{w}^\tau_{\text{out}} = \mathbf{a}_\tau,
\end{equation}
which we can plug into the memory capacity to obtain
\begin{equation}
\begin{aligned}
M_\tau^* &= \dfrac{\text{cov}^2(r(t-\tau),\mathbf{a}_\tau\mathbf{x}(t))}{ \text{var}(\mathbf{a}_\tau\mathbf{x}(t))} \\
& = \dfrac{\text{cov}^2(r(t-\tau),\|\mathbf{a}_\tau\|^2 r(t-\tau) + \langle \mathbf{a}_\tau,\boldsymbol{\varepsilon}_r(t) \rangle)}{
	\text{var}(\|\mathbf{a}_\tau\|^2 r(t) + \langle \mathbf{a}_\tau,\boldsymbol{\varepsilon}_r(t) )}
\end{aligned}
\end{equation}
where $M_\tau^*$ is the maximum memory that can be achieved by shifting the linear side of Eq.~\ref{eq:linearViewOnRC} but maintaining the linear-nonlinear ratio of variance. We must recall the definition of $\boldsymbol{\varepsilon}_r(t)$ in Eq.~\ref{eq:linearViewOnRC}, which implies that any correlation between its projection on $\mathbf{a}_\tau$ and $r(t)$ would imply that part of $r(t-\tau)$ is projected linearly onto the non-linear component. This necessarily means that $\text{cov}(\langle r(t-\tau),\langle \mathbf{a}_\tau,\boldsymbol{\varepsilon}_r(t)\rangle ) = 0$,
hence our previous equation becomes
\begin{equation}
M^*_\tau= \dfrac{\|\mathbf{a}_\tau\|^2}{\|\mathbf{a}_\tau\|^2 + \text{var}\left(\langle\frac{\mathbf{a}_\tau}{\|\mathbf{a}_\tau\|},\boldsymbol{\varepsilon}_r(t)\rangle\right) },
\end{equation}
and for the sake of simplicity we will name $\|\mathbf{a}_\tau\|^2 =a_\tau$, and $\text{var}\left(\langle \frac{\mathbf{a}_\tau}{\|\mathbf{a}_\tau\|}, \boldsymbol{\varepsilon}_r(t) \rangle \right) = \varepsilon_\tau$, leaving us with 
\begin{equation}\label{eq:MemoryBound}
M^*_\tau= \dfrac{a_\tau}{a_\tau + \varepsilon_\tau},
\end{equation}
where $a_\tau$ is the squared modulus of the linear projection of the input $r(t-\tau)$ on the reservoir state and $\varepsilon_\tau$ is the variance in that direction induced by the non-linear terms in that same direction.  

Hence the new problem that we must solve is to maximize 
\begin{equation}
\sum_{\tau = 1}^{N} M_\tau^* = \sum_{\tau = 1}^{N} \dfrac{a_\tau}{a_\tau + \varepsilon_\tau}
\end{equation}
subject to the constraint
\begin{equation}\label{eq:constraintMem1}
\sum_{\tau=1}^{N} a_\tau = c,
\end{equation}
with an order on the coefficients arising from the contractiveness or the reservoir
\begin{equation}\label{eq:constraintMem2}
a_{\tau + 1} < a_{\tau}
\end{equation}
and under the assumption that we do not change the nonlinear effects on any direction, hence $\varepsilon_\tau$ will be fixed. 

Finally, we shall also note that we will assume that $M^*_\tau$ is always a monotonically decreasing variable that goes from $M^*_1\approx1$ to $M^*_N \approx 0$, as we observed in the plots on Fig.~\ref{fig:MemoryCurves}. By noting that
\begin{equation}
M_\tau^* = \dfrac{1}{1+\frac{\varepsilon_\tau}{a_\tau}},
\end{equation}
this assumption implies that $q_\tau = \frac{\varepsilon_\tau}{a_\tau}$ goes from $r_1\approx 0$ to $r_N \rightarrow \infty$, hence $\varepsilon_\tau$ starts being much smaller than $a_\tau$ and decreases much slower than $a_\tau$. 

\subsubsection{Optimizing the linear projections}

Now the problem is to find how a change in the distribution of $\left[a_1, a_2, ..., a_N\right]$ would affect the value of $M^*$. Our approach will be to show that the fastest $a_\tau$ decreases, the lower $M^*$. 

We will show that there is one case, namely $a_\tau \propto \varepsilon_\tau$ which has higher $M^*$ than an other setting where $a_\tau$ decreases faster. Since we know that the values of $a_\tau$ must decrease faster than $\varepsilon_\tau$, our best arrangement of the strength of the linear projection of $r(t)$ is to try to make $a_\tau$ decrease as slowly as possible.

With a fixed ratio $a_\tau = \chi \varepsilon_\tau$, the upper bound on the memory is
\begin{equation}
M^* = \sum_{\tau=1}^{N}\dfrac{a_\tau}{a_\tau + \varepsilon_\tau} 
= \sum_{\tau=1}^{N}\dfrac{1}{1 + \frac{\varepsilon_\tau}{a_\tau}} 
= N \dfrac{1}{1 + \chi}.
\end{equation}
Now we will split the sequence of $M_\tau^*$ into two subsequences, one with $\tau<k$  where the values of $a_\tau$ will be increased by a factor $\beta_+$ and another one with $\tau>k+1$ where the values will be decreased by $\beta_-$. This leads us to the new memory capacity bound,
\begin{equation}
M^* =  k\dfrac{\beta_+}{\beta_+ + \chi} + (N-k) \sum_{\tau = k+1}^{N} \dfrac{\beta_-}{\beta_- + \chi} 
\end{equation}
which for simplicity we will normalize to obtain
\begin{equation}
m^* = \dfrac{M^*}{N} 
= \phi\dfrac{\beta_+}{\beta_+ + \chi} + (1-\phi) \dfrac{\beta_-}{\beta_- + \chi}.
\end{equation}
where $\phi = \frac{k}{N}$, and $m^*$ is just a normalized memory capacity bound.

Note that the function $m^*$ is still subject to the constraints presented in Eq.~\ref{eq:constraintMem1} and Eq.~\ref{eq:constraintMem2}. By introducing $\gamma = \frac{1}{c}\sum_{\tau=1}^{k} a_\tau $ we obtain
\begin{equation}\label{eq:constraintMem3}
\beta_+ \gamma + \beta_- (1-\gamma) = 1.
\end{equation}

Now we can introduce another variable $q = \frac{\beta_+}{\beta_-}$ which determines the difference between $\beta_+$ and $\beta_-$. As our main point is to show that a faster decrease of $a_\tau$ would decrease $m^*$, which in this particular case translates to
\begin{equation}
\dfrac{\partial m^*}{\partial q} < 0,
\end{equation}
which would imply that $\beta_+$ is larger than $\beta_-$ and hence $a_\tau$ decreases faster than $\varepsilon_\tau$ at $\tau=k$ and equally at any other $\tau$.

We can now compute the derivative
\begin{equation}
\begin{aligned}
\dfrac{\partial m^*}{\partial q} 
&= \phi\dfrac{\chi}{(\beta_+ + \chi)^2}  \dfrac{\partial \beta_+}{\partial q}
+ (1-\phi)\dfrac{\chi}{(\beta_- + \chi)^2}  \dfrac{\partial \beta_-}{\partial q}
\end{aligned}
\end{equation}
where the first term is positive and the second is negative -- because the memory capacity for $\tau<k$ will increase when $\beta_+$ grows and vice-versa. Since $\beta_+ > 1 > \beta_-$, $\frac{\chi}{(\beta_+ + \chi)^2}  < \frac{\chi}{(\beta_- + \chi)^2}$ we only need to prove that
\begin{equation}\label{eq:inequalityMem}
\phi \dfrac{\partial \beta_+}{\partial q} < - (1-\phi) \dfrac{\partial \beta_-}{\partial q}.
\end{equation}
To do so we will use the constraint from Eq.~\ref{eq:constraintMem3}. By setting $\beta_+ = \beta_-q$ we can solve both equations and obtain
\begin{equation}
\begin{aligned}
\beta_+ &= \dfrac{1}{\gamma}\dfrac{q}{q + \frac{1-\gamma}{\gamma}}
\Rightarrow \dfrac{\partial \beta_+}{\partial q} &= \dfrac{\frac{1- \gamma}{\gamma}}{(q+\frac{1-\gamma}{\gamma})^2}\\
\beta_- &= \dfrac{1}{\gamma}\dfrac{1}{q+\frac{1-\gamma}{\gamma}}
\Rightarrow \dfrac{\partial \beta_-}{\partial q} &= -\dfrac{1}{(q+\frac{1-\gamma}{\gamma})^2}.
\end{aligned}
\end{equation}
We can plug this into Eq.~\ref{eq:inequalityMem}, which simplifies to 
\begin{equation}
\begin{aligned}
&\dfrac{\phi}{\gamma} \dfrac{\frac{1-\gamma}{\gamma}}{(q+\frac{1-\gamma}{\gamma})^2} < \dfrac{1-\phi}{\gamma_+}\dfrac{1}{(q+\frac{1-\gamma}{\gamma})^2} 
\\
&\iff \phi (1-\gamma) < (1-\phi)\gamma \iff \phi < \gamma,
\end{aligned}
\end{equation}
which is true by the definitions of $\gamma$ and $\phi$, and the fact that $a_\tau$ is decreasing.

Now we have proven that whenever we can take an index $k$ and increase $a_\tau$ for $\tau > k$ and decrease $a_\tau$ for $\tau<k$, the memory capacity bound $M^*$ decreases. We shall now use this result to prove that whenever
\begin{equation}\label{eq:constraintMem4}
\dfrac{\epsilon_{\tau+i}}{\epsilon_{\tau}} < \dfrac{a_{\tau+i}}{a_\tau}\quad \forall i,\tau > 0,
\end{equation}
the memory capacity bound $M^*$ is lower than when $\varepsilon_\tau\propto a_\tau$. 

We do so by starting with the case $\varepsilon_\tau\propto a_\tau$ and we will change it step by step to a series of $a_\tau'$ that fulfills Eq.~\ref{eq:constraintMem4}. This is a simple iterative procedure.
We start with $k=1$, then select $q=\frac{\beta_+}{\beta_-}$ such that $a_k\beta_+ = a_k'$ under the constraint from Eq.~\ref{eq:constraintMem3}. Then we fix $a_1$ and we have the same problem for $a_\tau$ starting at $\tau\geq2$. This process can be repeated until $k=N$, at which point the memory bound $m^*$ will be lower because every modification with index $k$ lowered it.

Note that we have used $a_\tau$ in our argument, hence we obtained that the distribution of $a_\tau$ should be as homogeneous as possible. However, our result can also be stated for $a_\tau+\varepsilon_\tau$, because the non-linear effects of the reservoir on every direction are unchanged and the series $\varepsilon_\tau$ is also a decreasing one.

\subsubsection{Correlations as constraints on the variance}

From the previous section we know that the memory bound increases when the variance along the projections of the input into the reservoir state become more homogeneous. This can be expressed in terms of the state space of the reservoir. Intuitively, the variances at directions $\mathbf{a}_\tau$ must fit into the variances of the state space, and since we already established orthogonality of the projections, those variances must be along orthogonal directions. Since our goal is to have a variance as homogeneous as possible along the directions of $\mathbf{a}_\tau$, we need variance that is as homogeneous along orthogonal directions. Finally, this homogeneity is reduced when we add correlations between neurons.

We shall recall that we started our discussion by assuming that the probability distribution of neuron states is unchanged, which would ensure that the strength of the nonlinearity is not altered. This implies that the distribution of variances of the neurons is fixed. If we start by having zero correlations, we can start by setting 
\begin{equation}
a_\tau+\varepsilon_\tau = \text{var}\left(x_{\text{sort}(\tau)}(t)\right)
\end{equation}
where $\text{sort}(\tau)$ is the operation that finds the neuron with the $\tau$th largest variance in the reservoir. In other words, we associate every neuron to one direction of $\mathbf{a}_\tau$, with the constraint that the variances along those directions are ordered, hence we associate $\mathbf{a}_1$ to the neuron with highest variance, $\mathbf{a}_2$ to the second and so on. 

The distribution of $a_\tau+\varepsilon_\tau$ in that particular case is then given by the distribution of $\text{var}\left(x_{n}(t)\right)$. If the correlations are not zero, however, we need a new family of vectors which preserves orthogonality across the covariance matrix $\mathbf{C}$. This is given by the eigenvectors of $\mathbf{C}$. In that framework, the new variances are given by the eigenvalues of the covariance matrix, $\lambda_n(\mathbf{C})$. Naturally, when the correlations are zero, the eigenvalues correspond to the entries of the diagonal, which in our case are the variances as in the previous case. 

Hence we have to now work on the distribution of the eigenvalues of the covariance matrix. Specifically, we would want to show that increasing the correlations between neurons increases the inhomogeneity of the eigenvalues, which would decrease our memory bound $M^*$. A simple way to quantify this inhomogeneity is the mean with respect to the square root of the raw variance, which is given by
\begin{equation}
\nu = \dfrac{\sum_{n=1}^{N} \lambda_n^2(\mathbf{C})}
{\left(\sum_{n=1}^{N} \lambda_n(\mathbf{C}) \right)^2 },
\end{equation}
where $\lambda_n(\mathbf{C})$ is the $n$th eigenvalue of $\mathbf{C}$. To get an intuition of how this metric reflects the inhomogeneity, consider the case of two eigenvalues $\lambda_1,\ \lambda_2$; when $\lambda_1 = \lambda_2$ --very homogeneous -- then $\nu = \frac{1}{2}$, but when $\lambda_1 > 0, \lambda_2 = 0$ -- very inhomogeneous--, then  $\nu = 1$. For $N\gg1$ the perfectly homogeneous case approach zero but the perfectly inhomogeneous one is still one.

We can compute $\left(\sum_{n=1}^{N} \lambda_n(\mathbf{C}) \right)^2 $ by using th relationship between trace and eigenvalues,
\begin{equation}
\sum_{n=1}^{N} \lambda_n(\mathbf{C}) = \text{tr}\left[\mathbf{C}\right]
= \sum_{n=1}^{n} \text{var}\left(x_n(t)\right)
\end{equation}
which is constant by the assumption that the probability distributions of the neuron activities are fixed. Hence we can focus on the value of $\sum_{n=1}^{N} \lambda_n^2(\mathbf{C})$. This is easily done by noting that 
\begin{equation}
\mathbf{C}^kv_n(\mathbf{C}) = \lambda_n(\mathbf{C})\mathbf{C}^{k-1}v_n(\mathbf{C})
= \lambda_n^k(\mathbf{C})v_n(\mathbf{C})
\end{equation}
where $v_n(\mathbf{C})$ and $\lambda_n(\mathbf{C})$ are, respectively the $n$th eigenvector and eigenvalue of $\mathbf{C}$. If we plug this into the relationship between trace and eigenvalues we obtain
\begin{equation}
\sum_{n=1}^{N} \lambda_n^2(\mathbf{C}) = \text{tr}\left[\mathbf{C}^2\right],
\end{equation}
which we can compute by decomposing the square of covariance matrix and obtain
\begin{equation}
\sum_{n=1}^{N} \lambda_n^2(\mathbf{C}) 
= \sum_{n=1}^{N}\sum_{m=1}^{N} \mathbf{C}_{nm}\mathbf{C}_{mn}
= \sum_{n=1}^{N} \text{cov}\left(x_n(t), x_m(t)\right)^2.
\end{equation}
Which obviously grows when the neurons become correlated. Hence, the inhomogeneity measured by $\nu$ grows.

\subsubsection{Example}

The bound developed in previous sections might seem a bit artificial and far from the standard practice of reservoir computing, particularly the notion that we can adapt the linear part of the dynamics as we want. To make it more understandable and to show that the bound is indeed sharp we will present a simple example where all our assumptions are easily verified and the bounds are sharp.

For this we consider a line of neurons with the input on the first one. That is, 
\begin{equation}
\mathbf{W}_{ij} = 
\begin{cases}
w \quad \iff j=i+1\\
0\quad \text{otherwise},
\end{cases}
\end{equation}
with $w<1$ to keep the contractivity and the input is only sent to the first neuron, meaning that $\mathbf{w}_{\text{in}} = \left[w_{\text{in}}, 0, 0,...\right]$. If we let $w_{\text{in}} \ll 1$, then the network is effectively linear, because the hyperbolic tangent is almost an identity around $0$. Then the reservoir state becomes
\begin{equation}
\mathbf{x}(t) \approx w_{\text{in}}\left[u(t), wu(t-1),..., w^Nu(t-N)\right]
\end{equation}
which corresponds to the case where $\varepsilon_\tau\approx0$, and the previous input can be easily recovered, and this gives us $M=M^* = N$, which is the memory maximum \cite{dambre2012information,jaeger2001short}. If we increase the value of $w_{\text{in}}$, the reservoir is described as
\begin{equation}
\mathbf{x}(t) = \left[\tanh\left(w_{\text{in}}u(t)\right), \tanh\left(w\tanh\left(w_{\text{in}}u(t-1)\right)\right),...\right]
\end{equation}
where a readout can be obtained for each delay but the nonlinearity of repeatedly applying the hyperbolic tangent makes the memory harder and harder to recover, so the factor $\frac{\varepsilon_\tau}{a_\tau}$ grows. Naturally, here $M=M^*$.

In either case, the covariance matrix is diagonal because the inputs $r(t),\ r(t-\tau)$ are uncorrelated. Notice that, if we add connections between neurons or if we feed inputs with some alternative $\mathbf{w}_{\text{in}}$, then we would be increasing the correlations because there would be more neurons feed by the same input. This can only decrease the memory, because a reservoir with a line of neurons achieves its maximum memory capacity \cite{dambre2012information,white2004short}.

\section{Correlations and eigenvalues in dynamical systems}\label{SI:CorrVsEig}

We will show that the larger the eigenvalues of $\mathbf{W}$, the lower the correlations. 

To do so we will first linearize the system presented in Eq.~\ref{eq:ESNModel}, which gives us 
\begin{equation}\label{eq:LinearDynamicalSystem}
\mathbf{x}(t) = \mathbf{W}\mathbf{x}(t-1) + \mathbf{w}^{\text{in}}u(t).
\end{equation}
This linearizion might seem unjustified, as a key requirement of a reservoir is that it must be non-linear \cite{jaeger2002tutorial} to provide the necessary diversity of computations that a practical ESN requires. However, here we are interested in the memory capacity, which is maximized for linear reservoirs \cite{jaeger2002tutorial,white2004short}. That is, by studying a linear system we are implicitly deriving an upper bound on the memory, similarly to the approach taken in the control-theoretical study of the effect of the spectral radius \cite{jaeger2001short}. Finally, note that this linearizion is within the parameters of the ESN from Eq.\ref{eq:ESNModel}, as it would suffice to set $\|\mathbf{w}^{\text{in}}\|\ll 1$.

Given that our system is linear, we can formulate the state of a single neuron $x_i(t)$ as
\begin{equation}\label{eq:neuronAsScalarProd}
x_i(t) = \sum_{k=0}^{\infty} \left(W^k \mathbf{w}^{\text{in}}\right)_i u(t-k) 
= \sum_{k=0}^{\infty} a_{i,k} u(t-k) = \langle\mathbf{a}_i,\mathbf{u}_t\rangle 
\end{equation}
where the vector $\mathbf{a}_i = \left[a_{i,0}, a_{i,1},...\right]$ represents the coefficients that the previous inputs $\mathbf{u}_t = \left[u(t), u(t-1),...\right]$ have on $x_i(t)$. We can then plug this into the covariance between two neurons, 
\begin{equation}
\begin{aligned}
\text{cov}\left(x_i,x_j\right) &=
\lim\limits_{T\rightarrow\infty}\frac{1}{T}\sum_{q=t}^{t+T} \langle\mathbf{a}_i,\mathbf{u}_{q}\rangle \langle\mathbf{a}_j,\mathbf{u}_{q}\rangle \\
&= \langle \mathbf{a}_i, \mathbf{a}_j\rangle  \lim\limits_{T\rightarrow\infty}\frac{1}{T}\sum_{q_i=0}^{T} \sum_{q_j=0}^{T} \langle\mathbf{u}_{q_i},\mathbf{u}_{q_j}\rangle,
\end{aligned}
\end{equation}
and given that $u(t)$ is a random time series with zero autocorrelation and variance of one,
\begin{equation}
\lim\limits_{T\rightarrow\infty}\frac{1}{T}\sum_{q_i=0}^{T} \sum_{q_j=0}^{T} \langle\mathbf{u}_{q_i},\mathbf{u}_{q_j}\rangle =  \lim\limits_{T\rightarrow\infty}\frac{1}{T}\sum_{q=0}^{T}  \langle\mathbf{u}_{q},\mathbf{u}_{q}\rangle = \mathbb{E}\left[u^2(t)\right] = 1.
\end{equation} 
This gives us
\begin{equation}\label{eq:CovarianceInnerProd}
\text{cov}\left(x_i,x_j\right) = \langle \mathbf{a}_i, \mathbf{a}_j\rangle,
\end{equation}
and similarly, we can compute the variance of $x_i$, 
\begin{equation}
\begin{aligned}
\text{var}\left(x_i\right)= \text{cov}\left(x_i,x_i\right) 
= \langle \mathbf{a}_i, \mathbf{a}_i  \rangle 
=\| \mathbf{a}_i\|^2.
\end{aligned}
\end{equation}
We can plug the previous two formulas into the Pearson's correlation coefficient between two nodes $i,j$ as:
\begin{equation}\label{eq:geometricCorrelation}
P_{ij} =
\dfrac{\langle \mathbf{a}_i, \mathbf{a}_j \rangle}
{\|\mathbf{a}_i\|\|\mathbf{a}_j\|} 
= \cos(\mathbf{a}_i,\mathbf{a}_j)
\end{equation}
which is the same as the cosine distance between vectors $\mathbf{a}_i$ and $\mathbf{a}_j$.

The next step is thus to write $\mathbf{a}_i$ as a function of the eigenvalues of $\mathbf{W}$. To do so, we note that the state of a neuron can be written as
\begin{equation}
\mathbf{x}(t) = \sum_{k=0}^{\infty} \mathbf{W}^k\mathbf{w}^{\text{in}}u(t-k) = \sum_{k=0}^{\infty} \left(V \mathbf{\Lambda}^k V^{-1}\right)\mathbf{w}^{\text{in}} u(t-k)
\end{equation}
where $V$ is the matrix eigenvectors of $\mathbf{W}$ and $\mathbf{\Lambda}$ the diagonal matrix containing the eigenvalues of $\mathbf{W}$. When we obtain
\begin{equation}
x_i(t)= \sum_{k=0}^{\infty} \sum_{n=1}^{N} \lambda_n^k \langle v^{-1}_n,\mathbf{w}^{\text{in}}\rangle (v_n)_i u(t-k) 
,
\end{equation}
where $v_n$ and $v^{-1}_n$ are, respectively, the left and right eigenvectors of $\mathbf{W}$. Notice that as long as the network given by $\mathbf{W}$ is drawn from an edge-symmetric probability distribution -- meaning that $\text{Pr}\left[\mathbf{W}_{ij} = a\right] = \text{Pr}\left[\mathbf{W}_{ji} = a\right]\ \forall a\in \mathbb{R}$-- and is self averaging then $v_n$ and $v^{-1}_n$ are vectors drawn from the same distribution. 

The $\lambda_n^k$ terms present in the previous equation can be used as a new vector basis, 
\begin{equation}
\begin{aligned}
x_i(t)=  \sum_{n=1}^{N} \langle v^{-1}_n,\mathbf{w}^{\text{in}}\rangle (v_n)_i \langle\boldsymbol{\lambda}_n, \mathbf{u}_t\rangle = \sum_{k=0}^{\infty} \sum_{n=1}^{N} \lambda_n^k b_{i,n} u(t-k),
\end{aligned}
\end{equation}
where $\boldsymbol{\lambda}_n = \left[1, \lambda_n, \lambda_n^2, ...\right]$ and $b_{i,n} = \langle v^{-1}_n,\mathbf{w}^{\text{in}}\rangle (v_n)_i$. By simple identification from Eq.~\ref{eq:neuronAsScalarProd}, we find that
\begin{equation}\label{eq:a_iasSumLambda}
\left(\mathbf{a}_i\right)_k = \sum_{n=1}^{N} \lambda_n^k b_{i,n}.
\end{equation}
Thus every coefficient of $\mathbf{a}_i$ is a sum of many terms. Specifically, every term is a multiplication of $b_{i,n}$, which are all independent as they refer to the projections of $v_{n,i}$ into $\mathbf{w_\text{in}}$ and the values of $\lambda_n$, whose phase -- which we assume to be uniformly distributed on $\left[0,2\pi\right]$-- ensures that $\left(\mathbf{a}_i\right)_k$ is uncorrelated with $\left(\mathbf{a}_i\right)_{k+1}$. 

We will now proceed to cast the distribution of $\mathbf{a}_i$ as a uniform distribution of points defining an ellipsoid. By the central limit theorem, the values of $\mathbf{a}_i$ are independent random variables drawn from a normal distribution with zero mean and whose variance decreases with the index $k$. Therefore the distribution of $\mathbf{a}_i$ is given by
\begin{equation}
\prod_{k=0}^{\infty} e^{-\frac{(\mathbf{a}_i)_k^2}{s^2_k}}
\end{equation}
where $s_k$ is a decreasing function of $k$. Thus all the points with probability $e^{-r^2}$ are given by the surface
\begin{equation}
\sum_{k=0}^{\infty} \dfrac{a_k^2}{s_k^2} - r^2 = 0
\end{equation}
which are ellipsoids of infinite dimension and axis $\frac{s_k}{r}$. Furthermore, as we saw on Eq.~\ref{eq:CorrelationCoeff}, we are only interested in the angular coordinates of the points in the ellipsoid, not on their distance to the origin. Thus we can project every one of those surfaces into an ellipsoid with axis
\begin{equation}
\boldsymbol{s} = \left[1,\frac{s_2}{s_1},\frac{s_3}{s_1},...\right].
\end{equation}  
Note that, even though the ellipsoid has infinite dimensions, the length of the axes decreases exponentially due to the factor $\lambda_n^k$. Therefore, it has finite surface and it can be approximated by an ellipsoid with finite dimensions.

Now we have that the vectors $\mathbf{a}_i$ are, ignoring their length, uniformly distributed on an ellipsoid with axis $\boldsymbol{\sigma}$. Then 
\begin{equation}\label{eq:SintegralEllipsoid}
\lim\limits_{N\rightarrow \infty}S = \dfrac{1}{2}\int_{E_{\boldsymbol{s}}} \int_{E_{\boldsymbol{s}}} \cos^2(\angle(p,q)) dp dq
\end{equation}
where the integrals are taken over $E_{\boldsymbol{s}}$, the ellipsoid with axes $\boldsymbol{s}$, and the half factor comes from counting every pair only once. 

If we now change to spherical coordinates we will find that the two vectors can be expressed as 
\begin{equation*}
	\begin{aligned}
		p &= \left[r_p, \phi^p_1, \phi^p_2,...\right]\\
		q &= \left[r_q, \phi^q_1, \phi^q_2,...\right],
	\end{aligned}
\end{equation*}
where $\phi^p_1$ is the angle of $p$ on the plane given by the first and second axis, $\phi^p_2$ the plane by the first and third axis and so on. The cosine between the two vector is then
\begin{equation}
\cos(\angle(p,q)) = \prod_{k=1}^{\infty}\cos(\phi^p_k - \phi^q_k).
\end{equation}
Thus we can write the integral from Eq.~\ref{eq:SintegralEllipsoid} as
\begin{equation}\label{eq:SasProdOfCos}
\lim\limits_{N\rightarrow \infty}S = \dfrac{1}{2}\prod_{k=1}^{\infty} \int_{0}^{2\pi}\cos^2\left(\phi^p_k - \phi^q_k\right) \mu_{\phi_k}(\phi^p_k) \mu_{\phi_k}(\phi^p_k)  d\phi^p_k d\phi^q_k
\end{equation}
where $\mu_{\phi_k}(\phi)$ is the probability density function of the difference angle $\phi^p_k - \phi^q_k$. 

We will show that this integral decreases when the values $s_k$ increase. To do so, it is helpful to consider the extreme cases to get an intuition: when the semi-minor axis is zero, then we have a line, and all the points in the line have an angle between them either of zero or $\pi$, and thus a squared cosine of one. Conversely, when the semi-minor axis is maximal it equals the semi-major one and we have a circle, and the average squared cosine becomes $1/2$. Those are the two extreme values and thus the squared cosine decreases as the ellipse becomes more similar to a circle.

To make this argument more precise, we start by finding the density $\mu_{\phi_k}(\phi^p_k)$. This density is found by taking a segment of differential length $dl_S$ on the sphere of radius one and then compare it with the length covered in the ellipse $dl_E$. This gives us 
\begin{equation*}
	\begin{aligned}
		\mu_{\phi_k}(\phi) &\propto \dfrac{dl_E}{dl_S}\\
		&= \dfrac{\|\cos(\phi-d\phi) - \cos(\phi), s_k^2(\sin(\phi-d\phi) - \sin(\phi))\|}{\phi +d\phi - \phi}\\
		&= \sqrt{\sin^2(\phi) + a_k^2\cos^2(\phi)} = \sqrt{1 - (1-s_k^2)\cos^2(\phi)}.
	\end{aligned}
\end{equation*}
To fully evaluate the previous integral we would need to normalize $\mu_{\phi_k}$ and then evaluate the integral as a function of $s_k$. However, we would take a simpler approach  and note that $s_k$ controls the homogeneity of $\mu_{\phi_k}$: the larger $s_k$ is (within the interval $\left[0,1\right]$), the more the mass of probability is concentrated on the area around $\phi\sim 0$ and $\phi\sim \pi$.

Furthermore, we note that the squared cosine has the following periodicities
\begin{equation*}
	\cos^2(\theta) = \cos^2(\pi + \theta) = \cos^2(\pi - \theta) = \cos^2(2\pi -\theta),
\end{equation*}
thus, when we integrate over the angle $\phi$ we can take advantage of the four-fold symmetry and integrate only on the interval $\left[0,\pi/2\right]$. Thus we only need to study the integral
\begin{equation}
\int_{0}^{\frac{\pi}{2}} \cos^2\left(\phi^p_k - \phi^q_k\right) \mu_{\phi_k}(\phi^p_k) \mu_{\phi_k}(\phi^p_k)  d\phi^p_k d\phi^q_k,
\end{equation}
and by using $\sin^2(\theta)+\cos^2(\theta) = 1$, we can recast the previous integral as 
\begin{equation}
1 - \int_{0}^{\frac{\pi}{2}} \sin^2\left(\phi^p_k - \phi^q_k\right) \mu_{\phi_k}(\phi^p_k) \mu_{\phi_k}(\phi^p_k)  d\phi^p_k d\phi^q_k.
\end{equation}
In the interval $\phi\in \left[0,\pi/2\right]$ the squared sine can be seen a metric between two angles. Therefore the term 
\begin{equation}
\int_{0}^{\frac{\pi}{2}} \sin^2\left(\phi^p_k - \phi^q_k\right) \mu_{\phi_k}(\phi^p_k) \mu_{\phi_k}(\phi^p_k)  d\phi^p_k d\phi^q_k
\end{equation}
is nothing else than an average distance between the points which have a density given by $\mu_{\phi_k}$. Thus, the more the density is homogeneous, the larger the distance and vice-versa. Putting it all together, $s_k$ controls the homogeneity of $\mu_{\phi_k}$, and the homogeneity of $\mu_{\phi_k}$ controls the terms on Eq.~\ref{eq:SasProdOfCos}. Specifically, increasing $s_k$ decreases $S$.

The last thing to mention is that the values of $|\lambda_n|$ control $s_k$, as they give the variance to $(\mathbf{a}_i)_k$ in Eq.~\ref{eq:a_iasSumLambda}. Thus, the larger $|\lambda_n|$ the higher $s_k$ and the lower $S$. By the negative correlation between $M$ and $S$, increasing the values of $|\lambda_n|$ should increase the memory.

\section{Maximizing the Memory Capacity}\label{SI:MaximizeM}

Given that the memory capacity is one of the main characteristics of a reservoir, it is natural to ask whether our set-up offers some insight into how to maximize it. Obviously, the direct answer is to maximize $\langle\lambda\rangle$, so making every eigenvalue such that $|\lambda_n|=1$. 

\begin{figure*}
	\centering
	\subfloat{{\includegraphics[width=0.5\textwidth,trim={0cm 0 0cm 0},clip]{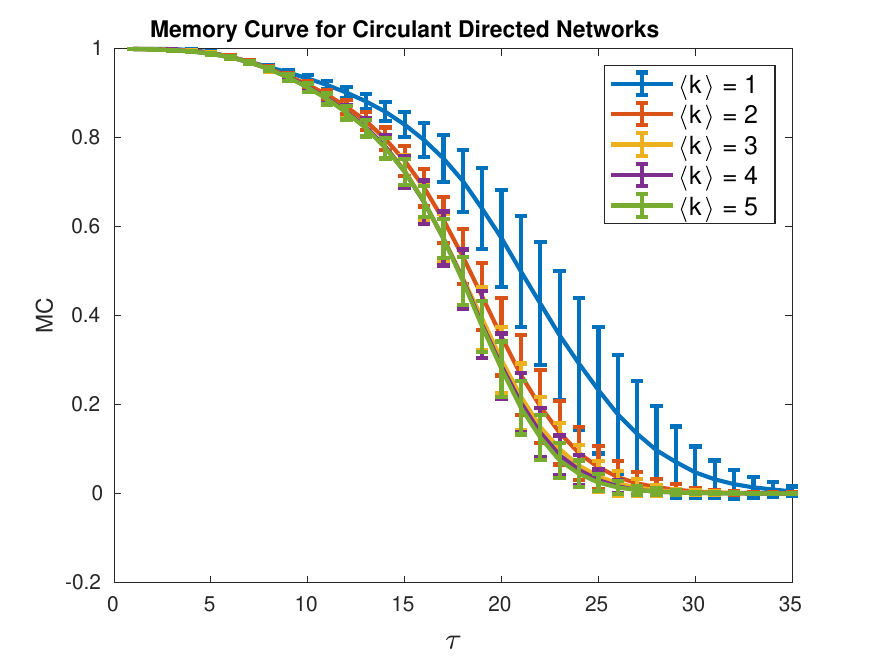}}}
	\subfloat{{\includegraphics[width=0.5\textwidth,trim={0cm 0 0cm 0},clip]{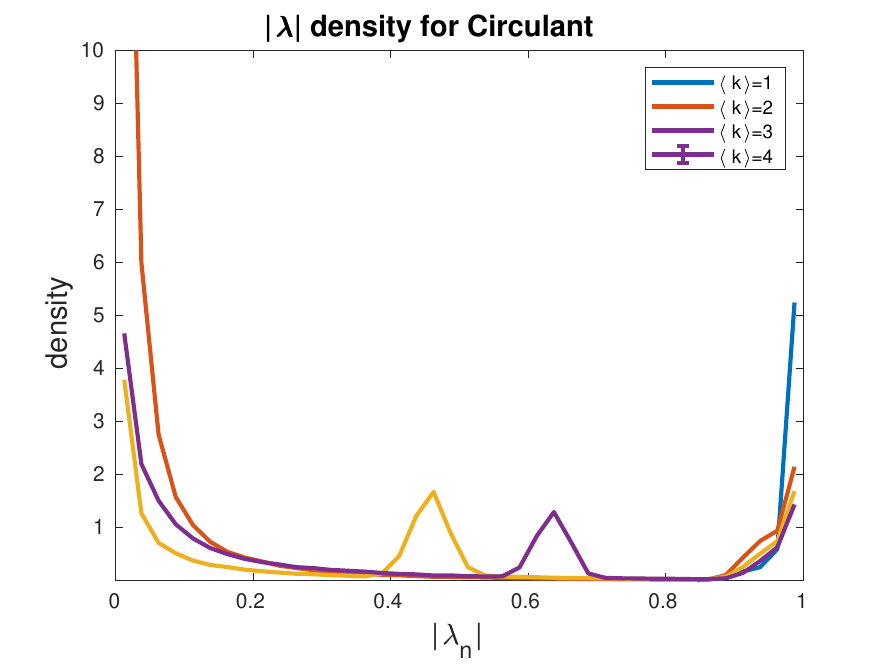}}}
	\caption{ Memory decay (left) and eigenvalue densities (right) for circulant networks with various degrees. The memory decays significantly slower for $\langle k\rangle=1$, corresponding to the eigenvalue distribution, which is all concentrated at $|\lambda| = 1$ for $\langle k\rangle=1$, but for  $\langle k\rangle>1$ more and more eigenvalues concentrate around lower $|\lambda|$.}
	\label{fig:MemoryCirculant}
\end{figure*}

There are many methods to create matrices with a given spectra, but they tend to be dense, thus we resort to directed circulant networks. Those networks have all nodes aligned in a circle and every node has a connection to the $d$ subsequent clockwise neighbors. For such a network with degree $d$, the corresponding is matrix such that
\begin{equation}
\mathbf{W}_{nm} = 
\begin{cases}
\mathcal{N}(0,1)\ \iff n-m \mod N \in \left[1,..,d\right]\\
0\ \text{otherwise}
\end{cases}
\end{equation}
which have eigenvalues distributed in $\lfloor\frac{d+1}{2}\rfloor$ concentric circles centered at the origin (see \cite{aceituno2018eigenvalues} for details). In any case, the eigenvalues for $d=1$ are the $N$th roots of the product of the weights, thus after rescaling we obtain $|\lambda_n|=1$. The resulting memory curve is presented in Fig.~\ref{fig:MemoryCirculant}

It is worth noticing that this particular architecture is very similar to having a long line of neurons, each one receiving inputs form its predecessor --the only difference being that the last neuron is coupled to the first. This line architecture can obviously provide the maximum memory $M=N$ is the input is scaled appropriately, but it is beyond the traditional rules of reservoir computing: its eigenvalues are aways zero, and its input weights need to be set as $\mathbf{w_\text{in}} = \left[s, 0, 0,...,0\right]$ with $s$ being almost zero. It is remarkable, however, that we obtain a similar architecture just by looking at $\langle\lambda\rangle$.

\section{Reservoir Design in the Fourier Domain}\label{SI:whyFreq}

The intuition that we will use here is that every neuron can be seen as a filter that extracts some features from the input time series. If the reservoir extract the right features, the ESN performance would improve. 

Training the readout through a linear regression is a minimization of the distance between the linear subspace spanned by the neurons' outputs and the target output, given by
\begin{equation}
\begin{aligned}
\|\mathbf{e}\|^2 &=  \sum_{t=t_0}^{T+t_0} (\hat{y}(t)- \mathbf{w}_{\text{out}} \mathbf{x}(t))^2 
&= \left\|\hat{y} - \sum_{n=1}^{N}r_n x_n\right\|^2
\end{aligned}
\end{equation} 
where $\|\cdot\|$ is the euclidean norm and $\mathbf{e}$ is the vector of errors, which inhabits the space of the time series. In this space, $\hat{y}$ is the target point, where every value of $\hat{y}(t)$ corresponds to the coordinate of $\hat{y}$ at dimension $t$. The time series of the neurons $x_n$ are also points in that space with $x_n(t)$ being their corresponding coordinates. Then, $\mathbf{w}_{\text{out}}\mathbf{x}$ is the linear combination of neuron points that is closest to $\hat{y}$. Having this geometrical interpretation of the ESN training it is clear that we should set $x_n$ to be as close as possible to the target $\hat{y}$, then the error should also be reduced. 

To make this statement more precise, we develop a bound that shows how the error in the training error changes as the neurons become more similar to the readout. We do so by decomposing every neuron time series 
\begin{equation}
x_n = x_n^{||y} + x_n^{\perp y}
\end{equation}
where $x_n^{||y}$ is the projection of $x_n$ onto the direction of $\hat{y}$ and $x_n^{\perp \hat{y}}$ the part that is orthogonal to it. We propose a very simple readout vector,
\begin{equation}
\left(\mathbf{w}_{\text{out}}^*\right)_n = s_n \gamma 
\end{equation}
where
\begin{equation}
\gamma = \dfrac{\| \hat{y}\|}{\sum_{n=1}^{N} \|x_n^{||\hat{y}}\|},\quad
s_n = \begin{cases}
1 \ \iff x_n^{||\hat{y}} > 0\\
-1 \ \iff x_n^{|| \hat{y}} \leq 0.
\end{cases}
\end{equation}
Thus the readout simply adds the vectors $x_n$ in the direction of $\hat{y}$ and scales them equally so that 
\begin{equation}\label{eq:orthogonalSubspace}
\sum_{n=1}^{N}\left(\mathbf{w}_{\text{out}}^*\right)_n x_n^{|| \hat{y}} = \hat{y}.
\end{equation}

Notice that our readout is obviously not the optimal, one, so the error that we get is \textit{harder},
\begin{equation}
\|e\| \leq \|e^*\| = \left\|\hat{y} - \sum_{n=1}^{N}\left(\mathbf{w}_{\text{out}}^*\right)_n x_n\right\|.
\end{equation}
We can now decompose $x_n$, obtaining
\begin{equation}
\left\|\hat{y}- \sum_{n=1}^{N} \left(\mathbf{w}_{\text{out}}^*\right)_n x_n\right\|^2 = \left\|\left(\hat{y}- \sum_{n=1}^{N} \left(\mathbf{w}_{\text{out}}^*\right)_n x_n^{|| \hat{y}}\right) + \sum_{n=1}^{N} r^*_n x_n^{\perp \hat{y}}\right\|^2,
\end{equation}
which, given Eq.~\ref{eq:orthogonalSubspace}, becomes
\begin{equation}
\|e^*\|^2 = \left\|\sum_{n=1}^{N} s_n x_n^{\perp \hat{y}}\right\|^2 
= \| \hat{y}\|^2\dfrac{\left\|\sum_{n=1}^{N} s_n x_n^{\perp \hat{y}}\right\|^2}
{\left(\sum_{n=1}^{N} \|x_n^{||\hat{y}}\|\right)^2}.
\end{equation}
In the worst case, $s_nx^{\perp \hat{y}}_n$ are all aligned on and on the same direction, so $\|\sum_{n=1}^{N} s_n x_n^{\perp y}\| = \sum_{n=1}^{N} \|x_n^{\perp \hat{y}}\|$. This yields the bound
\begin{equation}\label{eq:geometricBound}
\sigma^2 =  \dfrac{\| e\|^2}{\|\hat{y}\|^2} \leq  \dfrac{\| e^*\|^2}{\|\hat{y}\|^2} = \dfrac{\left(\sum_{n=1}^{N} \|x_n^{\perp \hat{y}}\|\right)^2}{\left(\sum_{n=1}^{N} \|x_n^{||\hat{y}}\|\right)^2}.
\end{equation}
The final step in this geometric bound is to notice that the basis that we used for our space is not unique; when we talk about $x_n^{\perp \hat{y}},\ x_n^{|| \hat{y}}, \hat{y}$, we do not need to use the basis where every entry corresponds to one time entry. We can instead choose another orthonormal basis and we will still keep the same values and distances. Our choice here is the Fourier basis, which is a natural choice when thinking about time series or signals \cite{elliott2013handbook}.
At this point it us useful to drop the geometric interpretation and come back to the time series or frequency formulation of $\hat{y}$ and $e^*$, giving
\begin{equation}\label{eq:FrequencyBoundGeometry}
\begin{aligned}
\sigma^2  \leq  \dfrac{\| e^*\|^2}{\|\hat{y}\|^2} = \dfrac{\left(\sum_{n=1}^{N} \|\mathcal{F}\left[x_n^{\perp \hat{y}}\right]\|\right)^2}{\left(\sum_{n=1}^{N}\|\mathcal{F}\left[x_n^{|| \hat{y}}\right]\|\right)^2},
\end{aligned}
\end{equation} 
where $\mathcal{F}\left[x_n\right]$ is the same as $x_n$ in the Fourier basis. By noting that 
\begin{equation}
\begin{aligned}
\|\mathcal{F}\left[x_n^{\perp \hat{y}}\right]\| &= | \mathcal{F}[x_n]\times \mathcal{F}[\hat{y}]|\ \|\hat{y}\|\\
\|\mathcal{F}\left[x_n^{|| \hat{y}}\right]\| &= | \langle \mathcal{F}[x_n], \mathcal{F}[\hat{y}] \rangle|\  \|\hat{y}\|\\
\end{aligned}
\end{equation}
we obtain the bound in Eq.~\ref{eq:FrequencyBound} of the main text.

\section{Generating Reservoirs with cycles}\label{SI:netWithCycles}

In order to generate networks with desired $\rho_L$ from Eq.~\ref{eq:rho}, we designed the following algorithm, which takes as parameters the number of neurons $N$; the connectivity $c$; $|\rho_L|\in\left[0,1\right]$, which is the portion of edges that are dedicated to cycles of length $L$ --the other ones being random--; and $s \in\{-1, 1\}$, which corresponds to the feedback sign.

\begin{description} 
	\item[If $L=1$:] $\quad $ Create a random sparse matrix $\mathbf{W}_\text{r}$ with $cN(N-1)$ non-zero entries. Normalize the spectral radius to 1 and then $\mathbf{W}\leftarrow \alpha\left((1-|\rho_1|)\mathbf{W}_\text{r}+s r_1 I\right)$, where $I$ is the identity matrix.
	\item[Else]\textbf{:}
	\item[  Step-1:]  $\ $ Create $\frac{|\rho_L| c N^2}{2L}$ permutations of $L$ numbers randomly picked from 1 to $N$ without replacement. Each permutation corresponds to $L$ nodes that will be connected form a cycle.
	\item[  Step-2:]  $\ $ For each cycle, draw the edge weights from $\mathcal{N}(0,1)$.
	\item[  Step-3:]   $\ $ For each cycle, if the sign of the product of the edge weights is not the same as $s$, multiply the last edge by $-1$. This gives the adjacency matrix $\mathbf{W}_\text{c}$.
	\item[  Step-4:]  $\ $ Create a random sparse matrix $\mathbf{W}_\text{r}$ with $\frac{(1-|\rho_L|) c N^2}{2}$ entries and weights drawn from  $\mathcal{N}(0,1)$. Then $\mathbf{W}=\left(\mathbf{W}_\text{r}+\mathbf{W}_\text{c}\right)$.
	\item[  Step-5:]  $\ $ Normalize to the desired average eigenvalue moduli $\langle |\lambda|\rangle$ by $\mathbf{W}\gets \frac{\mathbf{W}}{\frac{1}{N}\sum_{i=1}^N|\lambda_i(\mathbf{W})|}\langle |\lambda|\rangle$, where $\lambda_i(\mathbf{W})$ are the eigenvalues of $\mathbf{W}$
\end{description}

The special treatment of the case $L=1$ is due to the fact that with length of 1 if all edges are self loops the network is completely disconnected and the number of edges is at most $N$, meaning that for some values of $\rho_1$ the number of edges would be lower than the number required by the connectivity parameter.

\section{Adapting the Power Spectral Density in non-linear reservoirs}\label{SI:analyticFreq}

While the relationship between cycles and frequencies is very natural in linear systems \cite{elliott2013handbook}, we are dealing with a non-linear reservoir, meaning that we must prove that adding cycles does indeed modify the frequency response of the reservoir. This is a simple consequence of the monotonicity of the nonlinearity, which implies that adding a cycle of length $L$ will increase the autocorrelation with delay $L$ of the neurons embedded in the cycle, and this in turn increases the PSD of the neuron.

To make this more precise, we start by applying the Wiener-Khinshin theorem \cite{khintchine1934korrelationstheorie,wiener1930generalized} to a neuron indexed by $n$, 
\begin{equation}
\text{PSD}_{x_n}(f) = |\hat{x}_n(2\pi f)|^2 = \mathbb{E}_t\left[x_n(t)x_n(t-\tau)\right] = C_{x_n}(\tau),
\end{equation}
where $\frac{\tau}{T} = f$ and $C_{x_n}(\tau)$ is the autocorrelation function. Hence, to increase the PSD of neuron $n$ at frequency $f$ we need to increase the autocorrelation of neuron $n$ with itself at delay $fT$. To show that this can be done by rewiring the network so that there are many cycles of length $L$ with positive feedback, we start by writing the equation describing a neuron state,
\begin{equation}
x_n(t) = \tanh\left(v_n u(t) + \sum_{m=1}^N w_{mn}x_m(t-1)\right).
\end{equation}
By applying the mean value theorem
\begin{equation}
\begin{aligned}
&x_n(t) = g\left(x_n,t\right)
\left[v_n u(t) + \sum_{m\in \mathcal{S}_n}w_{mn}x_m(t-1)\right]\\
&g\left(x_n,t\right) = \tanh'
\left(c\left[v_n u(t) + \sum_{m\in \mathcal{S}_n}w_{mn}x_m(t-1)\right]\right)
\end{aligned}
\end{equation}
where $c_{t,n}\in \left[0,1\right]$. This can be expanded recursively to obtain
\begin{equation}\label{eq:meanValThmWithPath}
\begin{aligned}
x_n(t) = \sum_{l=1}^{\tau-1} \sum_{m=1}^{N} \sum_{\mathbf{p}\in \mathcal{P}_l(n,m)} G(\mathbf{p},t)w_\mathbf{p} v_m u(t-l)\\
+ \sum_{m=1}^{N} \sum_{ \mathbf{p}\in \mathcal{P}_\tau(n,m)} G(\mathbf{p},t)w_\mathbf{p} x_m(t-\tau)
\end{aligned}
\end{equation}
where $\mathcal{P}_l(n,m)$ are the paths from neuron $n$ to neuron $m$ with length $l$ such that each vector $\mathbf{p}=\left[n, p_1, p_2, ... ,p_{l-1}, m\right]$ contains the indexes of the neurons with $p_0=n$ and $p_{\dim(\mathbf{p})} = m$. The function $G(p,t) = \prod_{k=1}^{\dim(\mathbf{p})}g(p_k,t-k)$ corresponds to the attenuation given by the nonlinearity along each path, and finally, $w_p = \prod_{k=1}^{\dim(\mathbf{p})} w_{nm}$ is the cumulative weight of path $\mathbf{p}$.

Sparse random large graphs such as the ones used for ESN reservoirs are locally tree-like\cite{wormald1999models,bollobas2010handbook}, meaning that for $\tau \ll N$, almost all the neuron states $x_m(t-\tau)$ in the last term of Eq.~\ref{eq:meanValThmWithPath} are distinct and $n\neq m$. However, when we rewire our graph such that it has many cycles with length $\tau$, we obtain
\begin{equation}
x_n(t) = Q(n,t) + x_n(t-\tau)\sum_{ \mathbf{c}\in \mathcal{P}_\tau(n,n)} G(\mathbf{c},t)w_\mathbf{c} ,
\end{equation}
where $Q(n,t)$ is the term that includes the first summand of Eq.~\ref{eq:meanValThmWithPath} as well as all the paths from the second summand that are not cyclic. By putting this into the autocorrelation,
\begin{equation}
\begin{aligned}
\mathbb{E}_t&\left[x_n(t)x_n(t-\tau)\right] =\overline{Q}(n) + \mathbb{E}_t\left[x_n^2(t)\right]\sum_{ \mathbf{c}\in \mathcal{P}_\tau(n,n)}w_\mathbf{c}\overline{G}(\mathbf{c}),
\end{aligned}
\end{equation}
where $\overline{Q}(n) = \mathbb{E}_t\left[Q(n,t)\right]$ and $\overline{G}(\mathbf{c})=\mathbb{E}_t\left[G(\mathbf{c},t)\right] $. 
Given that all the entries of the weight matrix and the input vector are randomly sampled from probability distributions with mean zero, we would expect $Q(n,t)$ to also have mean zero across all neurons, so 
\begin{equation}
\mathbb{E}_n\left[C_{x_n}(\tau)\right] = 
\mathbb{E}_n\left[\text{var}\left[x_n(t)\right]\sum_{ \mathbf{c}\in \mathcal{P}_\tau(n,n)}w_\mathbf{c}\overline{G}(\mathbf{c})\right]
\end{equation}
Here it is worth noticing that $G(\mathbf{c},t)$ is a multiplication of $g(m,t)>0$, hence it is always positive just as the variance of $x_n(t)$. Thus the strength of the average PSD of the neuron the signs of $w_\mathbf{c}$ 

An important insight from this derivation presented here is that by using the mean value theorem we loose information about the actual value of the autocorrelation. That is, while in linear systems we know\textit{ how much }the frequencies will be modified, in non-linear systems we do not, having to conform ourselves with knowing which frequencies will be enhanced or dampened.

\section{Algorithm to adapt reservoirs}\label{SI:designFreq}

In the tasks described in the main text, different tasks require different reservoir parameters. Specifically, for a given maximum cycle length value $L$ there is one combination of $\rho_l,\ \forall l\leq L$ and a value of $\langle|\lambda|\rangle$, which optimizes the ESN performance. In this section we present the heuristic that we use to find those parameters.

The first step is to tune the memory for the current task. We take a very simple approach, where we simply try many spectral radii on the interval $\left[0,1\right]$ and pick the one that gives the best performance on a classical ER network. Once found, we calculate the corresponding $\langle|\lambda|\rangle$ and we will use that for the rest of the process.

Since the reservoirs that we use are not linear, we need to characterize their frequency response for various values of $\rho_l$ for all the $l\leq L$ considered. This response, that we denote $\hat{R}(\rho_l,L)$ is computed by generating Gaussian noise with the same variance and mean as the original signal and use it as an input for the reservoir. Then apply the Fast Fourier Transform to the neurons' states and average over all neurons. As the reservoirs are generated randomly, it is necessary to average those responses over multiple reservoir instances. 
For a given $L$ we use the following heuristic to find the optimal combination of cycles with size no greater than $L$. 

\begin{description}
	
	\item[Step-1:] Compute the Fourier transform of the input signal and keep the vector of absolute values $\hat{s}=[\hat{s}(0), \hat{s}(2), ... \hat{s}(f_S)]$, where $f_S$ is half the sampling frequency.
	
	\item[Step-2:] Compute the scalar product $\langle \hat{s}, \hat{R}(\rho_l,l)\rangle$ for all $\rho_l,l$, and select the $\rho_l$ that maximizes it for each $l$.
	
	\item[Step-3:] Test the performance of an ESN with the values of $\rho_l$ found in the previous step. If the performance is lower than in the default case of $\rho_l=0$, do not optimize with regard to that length.
	
	\item[Step-4:] For all values of $\rho_l$ where the cycle length is allowed and which fill the condition $\|\rho\|_1 \leq 1$, select the one that maximizes $\sum_l \langle \hat{s}, \hat{R}(\rho_l,l)\rangle$.
\end{description}

\vspace{1cm}
\noindent{\bf \textbf{Author Contributions.}} Y.-Y.L. conceived and designed the project. P.V.A. performed all the analytical calculations and empirical data analysis. P.V.A. and G.Y. performed extensive numerical simulations. All authors analyzed the results. P.V.A. and Y.-Y.L. wrote the manuscript. G.Y. edited the manuscript.

\noindent{\bf Competing Interests.} The authors declare that they have no competing financial interests. 

\noindent{\bf Correspondence.} Correspondence and requests for materials should be addressed to Y.-Y.L. (yyl@channing.harvard.edu). 

\noindent{\bf Code availability.} The code used in this work is available for download through github under the following link:\\
\url{https://github.com/pvili/EchoStateNetworks_NetworkAdaptation/tree/master}

\section*{Acknowledgment}
We thank Professor Herbert J{\"a}ger and Benjamin Liebald for valuable discussions. This work was partially supported by the ``Fundació Bancaria la Caixa". 

\bibliographystyle{alpha}
\bibliography{./ESNDraft}

\newcommand{\etalchar}[1]{$^{#1}$}
\begin{thebibliography}{HKAW89}

\bibitem[Ace18]{aceituno2018eigenvalues}
Pau~Vilimelis Aceituno.
\newblock Eigenvalues of random graphs with cycles.
\newblock {\em arXiv preprint arXiv:1804.04978}, 2018.

\bibitem[Bis06]{christopher2006pattern}
Christopher~M. Bishop.
\newblock {\em Pattern Recognition and Machine Learning (Information Science
  and Statistics)}.
\newblock Springer-Verlag New York, Inc., Secaucus, NJ, USA, 2006.

\bibitem[BKM10]{bollobas2010handbook}
B{\'e}la Bollob{\'a}s, Robert Kozma, and Dezso Miklos.
\newblock {\em Handbook of large-scale random networks}, volume~18.
\newblock Springer Science \& Business Media, 2010.

\bibitem[BLM{\etalchar{+}}06]{boccaletti2006complex}
Stefano Boccaletti, Vito Latora, Yamir Moreno, Martin Chavez, and D-U Hwang.
\newblock Complex networks: Structure and dynamics.
\newblock {\em Physics reports}, 424(4):175--308, 2006.

\bibitem[BOL{\etalchar{+}}12]{boedecker2012information}
Joschka Boedecker, Oliver Obst, Joseph~T Lizier, N~Michael Mayer, and Minoru
  Asada.
\newblock Information processing in echo state networks at the edge of chaos.
\newblock {\em Theory in Biosciences}, 131(3):205--213, 2012.

\bibitem[BSL10]{busing2010connectivity}
Lars B{\"u}sing, Benjamin Schrauwen, and Robert Legenstein.
\newblock Connectivity, dynamics, and memory in reservoir computing with binary
  and analog neurons.
\newblock {\em Neural Computation}, 22(5):1272--1311, 2010.

\bibitem[BSVS08]{buteneers2008real}
Pieter Buteneers, Benjamin Schrauwen, David Verstraeten, and Dirk Stroobandt.
\newblock Real-time epileptic seizure detection on intra-cranial rat data using
  reservoir computing.
\newblock In {\em International Conference on Neural Information Processing},
  pages 56--63. Springer, 2008.

\bibitem[BY06]{buehner2006tighter}
Michael Buehner and Peter Young.
\newblock A tighter bound for the echo state property.
\newblock {\em IEEE Transactions on Neural Networks}, 17(3):820--824, 2006.

\bibitem[CLL12]{cui2012architecture}
Hongyan Cui, Xiang Liu, and Lixiang Li.
\newblock The architecture of dynamic reservoir in the echo state network.
\newblock {\em Chaos: An Interdisciplinary Journal of Nonlinear Science},
  22(3):033127, 2012.

\bibitem[Cou10]{coulibaly2010reservoir}
Paulin Coulibaly.
\newblock Reservoir computing approach to great lakes water level forecasting.
\newblock {\em Journal of Hydrology}, 381(1):76--88, 2010.

\bibitem[DS12]{deihimi2012application}
Ali Deihimi and Hemen Showkati.
\newblock Application of echo state networks in short-term electric load
  forecasting.
\newblock {\em Energy}, 39(1):327--340, 2012.

\bibitem[DSS{\etalchar{+}}12]{duport2012all}
Fran{\c{c}}ois Duport, Bendix Schneider, Anteo Smerieri, Marc Haelterman, and
  Serge Massar.
\newblock All-optical reservoir computing.
\newblock {\em Optics Express}, 20(20):22783--22795, 2012.

\bibitem[DVSM12]{dambre2012information}
Joni Dambre, David Verstraeten, Benjamin Schrauwen, and Serge Massar.
\newblock Information processing capacity of dynamical systems.
\newblock {\em Scientific Reports}, 2, 2012.

\bibitem[DZ06]{deng2006complex}
Zhidong Deng and Yi~Zhang.
\newblock Complex systems modeling using scale-free highly-clustered echo state
  network.
\newblock In {\em International Joint Conference on Neural Networks (IJCNN)},
  pages 3128--3135. IEEE, 2006.

\bibitem[Ell13]{elliott2013handbook}
Douglas~F Elliott.
\newblock {\em Handbook of digital signal processing: engineering
  applications}.
\newblock Academic Press, 2013.

\bibitem[FBG16]{farkavs2016computational}
Igor Farka{\v{s}}, Radom{\'\i}r Bos{\'a}k, and Peter Gergel'.
\newblock Computational analysis of memory capacity in echo state networks.
\newblock {\em Neural Networks}, 83:109--120, 2016.

\bibitem[FL11]{ferreira2011comparing}
Aida~A Ferreira and Teresa~B Ludermir.
\newblock Comparing evolutionary methods for reservoir computing pre-training.
\newblock In {\em International Joint Conference on Neural Networks (IJCNN)},
  pages 283--290. IEEE, 2011.

\bibitem[GHS08]{ganguli2008memory}
Surya Ganguli, Dongsung Huh, and Haim Sompolinsky.
\newblock Memory traces in dynamical systems.
\newblock {\em Proceedings of the National Academy of Sciences},
  105(48):18970--18975, 2008.

\bibitem[GTJ12]{manjunath2012theory}
Manjunath Gandhi, Peter Ti{\~n}o, and Herbert Jaeger.
\newblock Theory of input driven dynamical systems.
\newblock In {\em European Symposium on Artificial Neural Networks,
  Computational Intelligence and Machine Learning}, pages 25--27, 2012.

\bibitem[HAW89]{hubner1989dimensions}
U~H{\"u}bner, NB~Abraham, and CO~Weiss.
\newblock Dimensions and entropies of chaotic intensity pulsations in a
  single-mode far-infrared nh 3 laser.
\newblock {\em Physical Review A}, 40(11):6354, 1989.

\bibitem[HB10]{hammami2010improved}
Nacereddine Hammami and Mouldi Bedda.
\newblock Improved tree model for arabic speech recognition.
\newblock In {\em International Conference on Computer Science and Information
  Technology (ICCSIT)}, volume~5, pages 521--526. IEEE, 2010.

\bibitem[HKAW89]{huebner1989problems}
U~Huebner, W~Klische, NB~Abraham, and CO~Weiss.
\newblock On problems encountered with dimension calculations.
\newblock In {\em Measures of Complexity and Chaos}, pages 133--136. Springer,
  1989.

\bibitem[Jae01a]{jaeger2001short}
Herbert Jaeger.
\newblock {\em Short term memory in echo state networks}.
\newblock GMD-Forschungszentrum Informationstechnik, 2001.

\bibitem[Jae01b]{jaeger2001echo}
Herbert Jaeger\vspace{0mm}.
\newblock The ``echo state'' approach to analysing and training recurrent
  neural networks -with an erratum note.
\newblock {\em German National Research Center for Information Technology GMD
  Technical Report}, 148:34, 2001.

\bibitem[Jae02]{jaeger2002tutorial}
Herbert Jaeger.
\newblock {\em Tutorial on training recurrent neural networks, covering {BPPT},
  {RTRL}, {EKF} and the ``Echo State Network'' approach}.
\newblock GMD-Forschungszentrum Informationstechnik, 2002.

\bibitem[Jae05]{jaeger2005reservoir}
Herbert Jaeger.
\newblock Reservoir riddles: Suggestions for echo state network research.
\newblock In {\em Proceedings of the International Joint Conference on Neural
  Networks}, volume~3, pages 1460--1462. IEEE, 2005.

\bibitem[Jae07]{jaeger2007discovering}
Herbert Jaeger.
\newblock Discovering multiscale dynamical features with hierarchical echo
  state networks.
\newblock {\em Jacobs University Bremen, Technical Reports}, 2007.

\bibitem[JBS08]{jiang2008supervised}
Fei Jiang, Hugues Berry, and Marc Schoenauer.
\newblock Supervised and evolutionary learning of echo state networks.
\newblock In {\em Parallel Problem Solving from Nature--PPSN X}, pages
  215--224. Springer, 2008.

\bibitem[JH04]{jaeger2004harnessing}
Herbert Jaeger and Harald Haas.
\newblock Harnessing nonlinearity: Predicting chaotic systems and saving energy
  in wireless communication.
\newblock {\em Science}, 304(5667):78--80, 2004.

\bibitem[JLPS07]{jaeger2007optimization}
Herbert Jaeger, Mantas Luko{\v{s}}evi{\v{c}}ius, Dan Popovici, and Udo Siewert.
\newblock Optimization and applications of echo state networks with
  leaky-integrator neurons.
\newblock {\em Neural Networks}, 20(3):335--352, 2007.

\bibitem[Khi34]{khintchine1934korrelationstheorie}
Alexander Khintchine.
\newblock Korrelationstheorie der station{\"a}ren stochastischen prozesse.
\newblock {\em Mathematische Annalen}, 109(1):604--615, 1934.

\bibitem[Lic13]{Lichman:2013}
M.~Lichman.
\newblock {UCI} machine learning repository, 2013.

\bibitem[Lie04]{liebald2004exploration}
Benjamin Liebald.
\newblock {\em Exploration of effects of different network topologies on the
  {ESN} signal crosscorrelation matrix spectrum}.
\newblock PhD thesis, University Bremen, 2004.

\bibitem[LJ07]{lukovsevicius2007overview}
Mantas Luko{\v{s}}evicius and Herbert Jaeger.
\newblock Overview of reservoir recipes.
\newblock {\em Jacobs University Bremen, Technical Reports}, 2007.

\bibitem[LJ09]{lukovsevivcius2009reservoir}
Mantas Luko{\v{s}}evi{\v{c}}Ius and Herbert Jaeger.
\newblock Reservoir computing approaches to recurrent neural network training.
\newblock {\em Computer Science Review}, 3(3):127--149, 2009.

\bibitem[LYS09]{lin2009short}
Xiaowei Lin, Zehong Yang, and Yixu Song.
\newblock Short-term stock price prediction based on echo state networks.
\newblock {\em Expert systems with applications}, 36(3):7313--7317, 2009.

\bibitem[Mer76]{mermelstein1976distance}
Paul Mermelstein.
\newblock Distance measures for speech recognition, psychological and
  instrumental.
\newblock {\em Pattern Recognition and Artificial Intelligence}, 116:374--388,
  1976.

\bibitem[MG77]{mackey1977oscillation}
Michael~C Mackey and Leon Glass.
\newblock Oscillation and chaos in physiological control systems.
\newblock {\em Science}, 197(4300):287--289, 1977.

\bibitem[MNM02]{maass2002real}
Wolfgang Maass, Thomas Natschl{\"a}ger, and Henry Markram.
\newblock Real-time computing without stable states: A new framework for neural
  computation based on perturbations.
\newblock {\em Neural Computation}, 14(11):2531--2560, 2002.

\bibitem[NS12]{newton2012neurally}
Michael~J Newton and Leslie~S Smith.
\newblock A neurally inspired musical instrument classification system based
  upon the sound onset.
\newblock {\em The Journal of the Acoustical Society of America},
  131(6):4785--4798, 2012.

\bibitem[OXP07]{ozturk2007analysis}
Mustafa~C Ozturk, Dongming Xu, and Jos{\'e}~C Pr{\'\i}ncipe.
\newblock Analysis and design of echo state networks.
\newblock {\em Neural Computation}, 19(1):111--138, 2007.

\bibitem[PAGH03]{ploger2003echo}
Paul~G Pl{\"o}ger, Adriana Arghir, Tobias G{\"u}nther, and Ramin Hosseiny.
\newblock Echo state networks for mobile robot modeling and control.
\newblock In {\em Robot Soccer World Cup}, pages 157--168. Springer, 2003.

\bibitem[Par06]{parseval1806memoire}
Marc-Antoine Parseval.
\newblock M{\'e}moire sur les s{\'e}ries et sur l’int{\'e}gration
  compl{\`e}te d’une {\'e}quation aux diff{\'e}rences partielles
  lin{\'e}aires du second ordre, {\`a} coefficients constants.
\newblock {\em M{\'e}moires pr{\'e}sent{\'e}s par divers savants, Academie des
  Sciences, Paris,(1)}, 1:638--648, 1806.

\bibitem[PHG{\etalchar{+}}18]{pathak2018model}
Jaideep Pathak, Brian Hunt, Michelle Girvan, Zhixin Lu, and Edward Ott.
\newblock Model-free prediction of large spatiotemporally chaotic systems from
  data: a reservoir computing approach.
\newblock {\em Physical review letters}, 120(2):024102, 2018.

\bibitem[PMB13]{pascanu2013difficulty}
Razvan Pascanu, Tomas Mikolov, and Yoshua Bengio.
\newblock On the difficulty of training recurrent neural networks.
\newblock {\em International Conference on Machine Learning}, 28:1310--1318,
  2013.

\bibitem[RIA17]{rodriguez2017optimal}
Nathaniel Rodriguez, Eduardo Izquierdo, and Yong-Yeol Ahn.
\newblock Optimal modularity and memory capacity of neural networks.
\newblock {\em arXiv preprint arXiv:1706.06511}, 2017.

\bibitem[RT12]{rodan2012simple}
Ali Rodan and Peter Ti{\v{n}}o.
\newblock Simple deterministically constructed cycle reservoirs with regular
  jumps.
\newblock {\em Neural Computation}, 24(7):1822--1852, 2012.

\bibitem[SWL12]{strauss2012design}
Tobias Strauss, Welf Wustlich, and Roger Labahn.
\newblock Design strategies for weight matrices of echo state networks.
\newblock {\em Neural Computation}, 24(12):3246--3276, 2012.

\bibitem[TBCC07]{tong2007learning}
Matthew~H Tong, Adam~D Bickett, Eric~M Christiansen, and Garrison~W Cottrell.
\newblock Learning grammatical structure with echo state networks.
\newblock {\em Neural Networks}, 20(3):424--432, 2007.

\bibitem[VLS{\etalchar{+}}10]{verplancke2010novel}
Thierry Verplancke, S~Looy, Kristof Steurbaut, Dominique Benoit, F~Turck,
  G~Moor, and Johan Decruyenaere.
\newblock A novel time series analysis approach for prediction of dialysis in
  critically ill patients using echo-state networks.
\newblock {\em BMC Medical Informatics and Decision Making}, 10(1):1, 2010.

\bibitem[Wie30]{wiener1930generalized}
Norbert Wiener.
\newblock Generalized harmonic analysis.
\newblock {\em Acta mathematica}, 55(1):117--258, 1930.

\bibitem[WLS04]{white2004short}
Olivia~L White, Daniel~D Lee, and Haim Sompolinsky.
\newblock Short-term memory in orthogonal neural networks.
\newblock {\em Physical Review Letters}, 92(14):148102, 2004.

\bibitem[Wor99]{wormald1999models}
Nicholas~C Wormald.
\newblock Models of random regular graphs.
\newblock {\em London Mathematical Society Lecture Note Series}, pages
  239--298, 1999.

\bibitem[WW05]{whitley2005complexity}
Darrell Whitley and Jean~Paul Watson.
\newblock Complexity theory and the no free lunch theorem.
\newblock In {\em Search Methodologies}, pages 317--339. Springer, 2005.

\bibitem[YJK12]{yildiz2012re}
Izzet~B Yildiz, Herbert Jaeger, and Stefan~J Kiebel.
\newblock Re-visiting the echo state property.
\newblock {\em Neural Networks}, 35:1--9, 2012.

\end{thebibliography}

\end{document}